\documentclass[12pt]{article}

    \usepackage{amssymb,amsmath} %allow math symbol
    \usepackage{bbm} %allow \mathbbm{} in math env.
    \usepackage{nicefrac}
    \usepackage{booktabs} %allow three-line-table,  \toprule, \midrule and \bottomrule
    \usepackage{longtable} %allow for long table
    \usepackage{bigstrut} %allow big strut command
    \usepackage{enumerate} %allow custom enumerate env.
    \usepackage{graphicx} %allow figure included
    \usepackage{subcaption} % for subfigure
    \usepackage{multirow} %allow table of multirow included
    \usepackage{xcolor} %allow color
    \usepackage{amsthm} %allow custom theorem environment
    \theoremstyle{plain}
        \newtheorem{assumption}{Assumption}
    \theoremstyle{remark}
        \newtheorem{remark}{Remark}
    \theoremstyle{definition}
        
        \newtheorem{theorem}{Theorem}
        
    \usepackage{geometry}  %set distance between body and margin
    \usepackage{hyperref}
    \usepackage{xr}
    \usepackage{natbib} %bib
    \usepackage[ruled, linesnumbered]{algorithm2e} % algorithm
\newcommand{\loss}{\operatorname{\mathit{\ell}}}

\newcommand{\trans}{\intercal}
\newcommand{\T}{\intercal}
\newcommand{\wh}{\widehat}

\newcommand{\A}{\mathcal{A}}
\renewcommand{\H}{\mathcal{H}}

\newcommand{\bX}{\mathbf{X}}
\newcommand{\bbX}{\mathbb{X}}

\newcommand{\Op}{O_{\mathbb{P}}}
\newcommand{\op}{o_{\mathbb{P}}}

\newcommand{\of}[1]{\left(#1\right)}

\newcommand{\blind}{1}

\begin{document}
\def\spacingset#1{\renewcommand{\baselinestretch}%
{#1}\small\normalsize} \spacingset{1}
%%%%title page %%%%
% \title{Transfer Learning for Spatial Autoregressive Models with Application to U.S. Presidential Election}

\if1\blind
{
  \title{\bf Transfer Learning for Spatial Autoregressive Models with Application to U.S. Presidential Election Prediction}
  \author{Hao Zeng\(^{1}\), Wei Zhong\(^{2}\) and Xingbai Xu\(^{2}\)
  % \thanks{
  %   % The authors gratefully acknowledge \textit{please remember to list all relevant funding sources in the unblinded version}
  %   This work is supported by National Key R\&D Program of China (2022YFA1003800), the National Natural Science Foundation of China(NNSFC) (71988101, 72073110, 72333001, 12231011), National Statistical Science Research Grants of China (Major Program 2022LD08), and the 111 Project (B13028).
  %   }\hspace{.2cm}
    \\
    \({}^{1}\)Department of Statistics and Data Science\\
    Southern University of Science and Technology \\
    \({}^{2}\)MOE Key Lab of Econometrics\\ 
    WISE and Department of Statistics and Data Science in SOE\\
    Paula and Gregory Chow Institute for Studies in Economics\\
    Xiamen University}
  \maketitle
} \fi

\if0\blind
{
  \bigskip
  \bigskip
  \bigskip
  \begin{center}
    {\LARGE\bf Transfer Learning for Spatial Autoregressive Models with Application to U.S. Presidential Election}
\end{center}
  \medskip
} \fi

% \author{\name Hao Zeng \email zenghao.acmail@gmail.com \\
%        \addr Department of Statistics and Data Science\\
%        Southern University of Science and Technology\\
%        Shenzhen, Guangdong, China
%        \AND
%        \name Wei Zhong \email wzhong@xmu.edu.cn \\
%        \addr MOE Key Lab of Econometrics, WISE and Department of Statistics and Data Science in SOE\\
%        Xiamen University\\
%        Xiamen, Fujian, China
%        \AND
%        \name Xingbai Xu \email xuxingbai@xmu.edu.cn\\
%        \addr 
%        MOE Key Lab of Econometrics, WISE and Department of Statistics and Data Science in SOE\\
%        Paula and Gregory Chow Institute for Studies in Economics\\
%        Xiamen University\\
%        Xiamen, Fujian, China
%        }

%%%%document body

\bigskip
\begin{abstract}%
It is important to incorporate spatial geographic information into U.S. presidential election analysis, especially for swing states. The state-level analysis also faces significant challenges of limited spatial data availability. To address the challenges of spatial dependence and small sample sizes in predicting U.S. presidential election results using spatially dependent data, we propose a novel transfer learning framework within the SAR model, called as tranSAR. Classical SAR model estimation often loses accuracy with small target data samples. Our framework enhances estimation and prediction by leveraging information from similar source data. We introduce a two-stage algorithm, consisting of a transferring stage and a debiasing stage, to estimate parameters and establish theoretical convergence rates for the estimators. Additionally, if the informative source data are unknown, we propose a transferable source detection algorithm using spatial residual bootstrap to maintain spatial dependence and derive its detection consistency. Simulation studies show our algorithm substantially improves the classical two-stage least squares estimator. We demonstrate our method’s effectiveness in predicting outcomes in U.S. presidential swing states, where it outperforms traditional methods. In addition, our tranSAR model predicts that the Democratic party will win the 2024 U.S. presidential election.
\end{abstract}

\noindent%
\vfill

% \begin{keywords}
    % Spatial autoregressive model,
    % transfer learning,
    % U.S. presidential election,
    % spatial data analysis, 
    % high dimensional analysis.
% \end{keywords}

\newpage
\spacingset{1.9} % DON'T change the spacing!
\section{Introduction}\label{sec: introduction}
In recent years, amounts of Geographic Information Science (GIS) data have been used in election campaigns~\citep{peters2004gis}. Geography is the common denominator of where voters live, how residents engage in the voting process, how officials manage elections, and where elections campaign runs. Based on population data and geographic information, election district boundaries are resettled to draft multiple proposed plans and demonstrate the effects of several approaches. Political campaigns make full use of geographic information and related tools to obtain underlying support from voters, especially the ones from swing states, which are pivotal in determining the outcome of elections. Swing states, also known as battleground states, are crucial because they do not consistently support one party, making their election outcomes unpredictable and highly influential in election results. Shaping voter preference in swing states is an essential problem, especially for prediction. Including spatial information in data analysis brings a novel insight into the prediction.
There are at least two challenges in using spatial information for predicting U.S. presidential elections: one is leveraging spatial information effectively to predict election outcomes with considering the impact of spatial factors, and the other is the limited data from swing states. Finding ways to use information from other states to analyze spatial factors and predict the final result is crucial to overcoming these challenges.

Spatial dependence among spatial units exists widely in various fields including economics \citep{anselin1988spatial}, finance \citep{grennan2019dividend},  political science \citep{peters2004gis}, ecology \citep{verhoef2018spatial} and climate sciences \citep{okunlola2021spatial} and so on. 
The spatial autoregressive (SAR) model \citep{cliff1973spatial,ord1975estimation,cliff1975model} and its variants are extensively applied in numerous empirical studies concerning spatial competition and spatial spillover effects. 
For instance, the SAR model is employed to investigate the crime rates in 49 areas of Columbus, Ohio~\citep{anselin1988spatial}, budget spillover effects and policy interdependence~\citep{case1993budget}, cigarette demands in various states in the U.S. \citep{baltagi2004prediction}, agricultural land prices in Northern Ireland \citep{kostov2009spatial}, the mutual influence on the stock returns of companies with the same headquarters location \citep{pirinsky2006does} and the group effect of the dividend policy of listed companies \citep{grennan2019dividend}.

When the sample size of the target data is limited, it is possible to obtain less precise estimates for most statistical models. This issue is particularly relevant in spatial econometrics, where sample sizes can often be quite limited. For instance, in the study investigating crime rates in Columbus, Ohio, the sample size is restricted to just 49~\citep{anselin1988spatial}. This limitation underscores the necessity for additional research aimed at leveraging information from similar datasets to elevate estimation performance. Consequently, there is a pressing need for the exploration of methods to improve the estimation performance of the SAR models through the incorporation of information derived from similar source datasets.
Transfer learning is a useful approach to  incorporate information from source datasets~\citep{pan2009survey, torrey2010transfer, zhuang2021comprehensive, hu2023optimal}. The main idea of transfer learning is that the knowledge acquired in one context can be transferred to improve the performance of estimation or prediction  when applied to another, even if the two tasks or domains are not identical. In transfer learning, there are typically two main components:
(1) The first one is the \textit{target domain}, to which the knowledge is transferred. The target domain is of our interest but particularly its  sample size is limited. 
(2) The second one is the \textit{source domain}, from which the knowledge is transferred. In general, the model is pre-trained using the source datasets.

In the recent literature, \citet{bastani2021predicting} introduced a two-step estimator that employs high-dimensional techniques to capture biases between the target and source data and has provided an upper bound for the proposed estimators.
\citet{li2022transfer} considered transfer learning for estimation and prediction of a high dimensional linear regression and proposed a novel data-driven Trans-Lasso method for detecting the transferable sources. 
\citet{tian2023transfer} introduced a novel entropy-based method for identifying transferable sources and further constructed confidence intervals for each coefficient under the generalized linear model.
For more reference on related works, see \citet{cai2019transfer}, \citet{cai2024transfera, cai2024transfer}, and \citet{li2022transfera, li2022estimation}.

In this paper, in order to deal with two aforementioned challenges, spatial dependence and sample limitation, in U.S. presidential election prediction problems, we present a transfer learning framework within the SAR model, which we refer to as \textbf{tranSAR}, designed to improve both estimation and prediction.
Our study makes three contributions from methodology, statistical theory and empirical application in the literature of the SAR models. 
First, in methodology, we introduce a two-stage algorithm ($\A$-TranSAR), consisting of a transferring stage and a debiasing stage, to estimate the unknown parameters when the informative source data indexed by $\A$ are known. If we do not know which sources to transfer, a transferable source detection algorithm is proposed to detect informative sources data based on spatial residual bootstrap to retain the spatial dependence. To the best of our knowledge, this is the first work to introduce the transfer learning to the SAR models.
Second, in statistical theory, we establish the theoretical convergence rates for the $\A$-TranSAR estimators under the SAR model and theoretically show that the transfer learning can improve the estimation accuracy with the help of the information from the transferable source data.
We also derive the detection consistency of the transferable source detection algorithm to avoid the negative transfer when which sources to transfer is unknown.
Third, in the empirical application, we apply our method to predict the election outcomes in swing states in the U.S. presidential election via utilizing polling data from the last U.S. presidential election along with other demographic and geographical data. The empirical results show that our method outperforms traditional spatial econometrics methods.

Here, we emphasize the differences between our work and \cite{tian2023transfer}. 
Although our work shares a similar idea of transfer leaning with the previous literature, there are at least two significant differences from both theoretical and algorithmic perspectives.
(1) \cite{tian2023transfer} is based on generalized linear models for independent data. In contrast, our research focuses on spatially dependent data, which brings significant challenges for theoretical guarantees. This is because previous theories heavily rely on concentration inequalities applicable to independent data, and few concentration inequalities have been established for spatially dependent data. To address this, we develope a new theoretical foundation tailored to spatially correlated data. 
(2) For the algorithmic aspect, the transferable detection method in \cite{tian2023transfer} relies on sample splitting to identify transferable sources, a technique that is inapplicable for spatial data as spatial data are not independent. To this end, we propose a new method based on spatial bootstrap and establish its theoretical guarantees.

In the rest of our paper, we firstly introduce the real application problem (predicting the U.S. presidential
election result) and the associated transfer learning framework for the SAR model in Section~\ref{sec: emp}. 
Then, we introduce its estimation and source detection algorithm in Section~\ref{sec: model}.
In Section~\ref{sec: simulation}, we evaluate the finite sample performance of our proposed method via simulation studies.
Further, the application to predict U.S. presidential elections with transfer learning is conducted in Section~\ref{sec: emp2}.
Section~\ref{sec: theory} establishes the estimation convergence rates of the transfer learning estimators for the SAR model and  derive the detection consistency of the transferable source detection algorithm.
We conclude the paper in Section~\ref{sec: conclusion}.
Additional simulation results, theoretical analysis, and all proofs are provided in the supplementary materials.

\paragraph{Notations.}
Let \([n] \equiv \{1, \ldots, n\}\).
For an index set \(\A \subseteq [K]\) and a sequence of positive integers \(\{n_k, k = 1, \dots, K\}\), \(n_{\A} \equiv \sum_{k\in \A} n_{k}\).
For a vector \(\mathbf{x} \in \mathbbm{R}^d\) and \(S\) being a subset of set \([d]\), \(\mathbf{x}_{S}\) is the sub-vector of \(\mathbf{x}\) with index \(S\).
The notation~\(a \lesssim b\) means there exists a positive constant \(C \text{ s.t. } a \leq C b\).
The notation~\(a \vee b\) means \(\max(a, b)\), and \(a \wedge b\) means \(\min(a,b)\).
The notation~\( a_n \asymp b_n\) means that \(a_n/b_n\) converges to a positive constant.
For any random sequences \(a_n\) and \(b_n\), \(b_n = O_{\mathbb{P}}(a_n)\) means \(\forall \epsilon >0, \exists M>0 \text{ s.t. } \limsup_n \mathbb{P}(|b_n| \geq M |a_n|) \leq \epsilon \), and \(a_n \ll b_n\) means \(|a_n| = \op(|b_n|)\), i.e.~\(\lim_{n \rightarrow \infty}\mathbb{P}(|\frac{a_n}{b_n}| \geq \varepsilon) = 0\) for any \(\varepsilon >0\).
For a vector \(\mathbf{x} = {(x_1, \dots, x_d)}^\trans \in \mathbbm{R}^d\), we define \(\|\mathbf{x}\|_1 = \sum_{j=1}^{d} |x_j|\), \(\|\mathbf{x}\|_2 = {(\sum_{j = 1}^{d} x_j^2)}^{1/2} \), and \(\|\mathbf{x}\|_\infty = \max_{1 \leq j \leq d} |x_j|\).
For two column vectors \(\mathbf{x}\) and \(\mathbf{y} \in \mathbbm{R}^d\), their inner product is \(\langle \mathbf{x}, \mathbf{y} \rangle = \mathbf{x}^\trans \mathbf{y}\).
For an \(n \times n\) matrix \(A\), we define \(\|A\|_F \equiv {(\sum_{i=1}^m \sum_{j=1}^n\left|a_{i j}\right|^2)}^{1 / 2}\), \(\|{A}\|_{\infty} \equiv  \max _i \sum_{j=1}^n\left|a_{i j}\right|\), its spectral norm \(\|A\|_2 \equiv\max _{x \in \mathbb{R}^n \backslash\{0\}} \frac{\|A x\|_2}{\|x\|_2}=\sqrt{\lambda_{\max}(A'A)}\), where \(\lambda_{\max}(A)\) is the maximum eigenvalue of \(A\) and \(\lambda_{\min}(A)\) is the minimum eigenvalue of \(A\).
For an \(n \times m\) matrix \(A\), \(A_{(i,j)}\) represents the element in its \(i\)-th row and \(j\)-th column. \(A_{(\cdot,j)}\) refers to the \(j\)-th column of \(A\), and \(A_{(i,\cdot)}\) represents the \(i\)-th row of matrix \(A\).

\section{Problem and Model Settings}\label{sec: emp}

In this work, we focus on predicting the U.S. presidential election results of each county in swing states by incorporating the spatial relationship among counties.
We try to predict the election results in 2020 and 2024 elections using the polls from the 2016 and 2020 election U.S. presidential election and some other demographic data and  geographic data, respectively. 
It is worth noting that swing states have undergone historical shifts over time. 
The process of identifying swing states in previous elections typically involves an assessment of the closeness of the vote margins in each state. 
This analysis incorporates historical election outcomes, opinion polls, political trends, recent developments since the previous election, and the specific attributes, strengths, or vulnerabilities of the candidates in contention.
% For the 2020 election, we select six swing states from mainstream journals and news media, including Florida, Georgia, Michigan, Minnesota, North Carolina, and Ohio.
% For 2024 election, we selected the following states as swing states for analysis, Arizona, Georgia, Florida, Pennsylvania, Michigan, Wisconsin, Ohio, and Iowa.
In Section~\ref{ssec: data}, we describe the motivation problem and data.  
Here, we reiterate the challenges we face, namely the utilization of spatial information and data scarcity. To this end, we propose a novel transfer learning framework based on the SAR model aimed at addressing above mentioned two challenges.

\subsection{Problem and Data}\label{ssec: data}
\begin{figure}[hbt!]%
    \centering
    \includegraphics[width=0.8\linewidth]{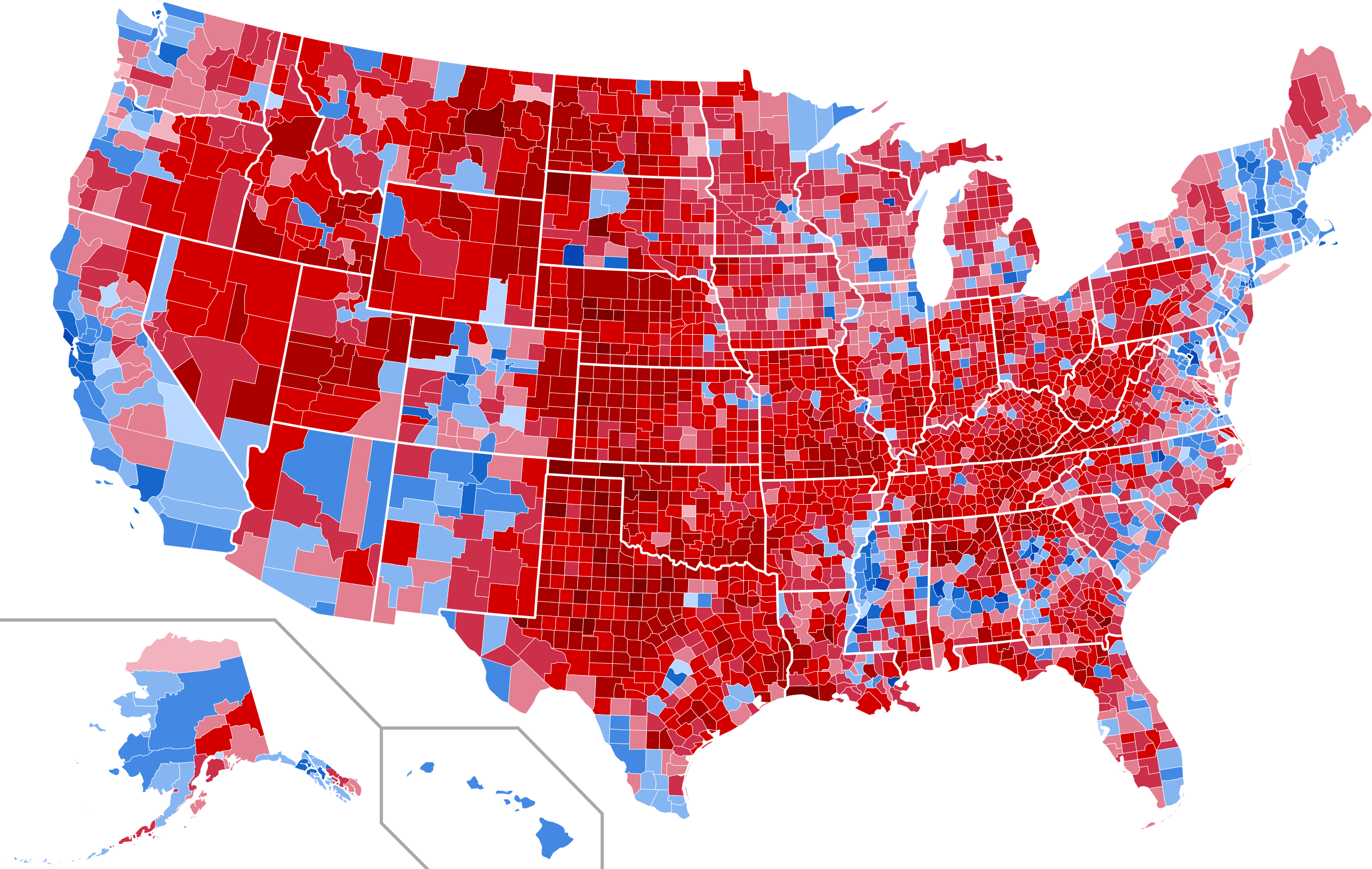}
    \caption{Results by county of the 2020 United States presidential election \citep{magogtheogre2020english}. The red areas represent support for Trump, while the blue areas indicate support for Biden. The intensity of the color corresponds to the degree of support.}
    \label{fig:map_US_2020}
\end{figure}

\paragraph{Problem:}
We aim to handle the challenges of utilizing spatial information and analyzing spatial factors in predicting the U.S. presidential election. From Figure~\ref{fig:map_US_2020}, it is evident that the county-level support rates in the U.S. presidential election display a spatial relationship. Spatial relationships among the data are also prevalent, making it essential to incorporate spatial information into the analysis.
In U.S.  presidential elections, the distribution of votes exhibits significant spatial dependency, a phenomenon particularly obvious over counties. Specifically, the voting patterns of a county tend to align closely with those of its neighboring counties. This similarity can often be attributed to the shared socio-economic characteristics that arise from geographical proximity, such as industrial structure, income levels, and educational attainment \citep{fotheringham2021scale, kim2003spatial}. Additionally, whether a county is urban or rural plays a substantial role in this spatial dependency. Urban counties generally show stronger support for the Democratic Party, whereas suburban and rural counties tend to favor the Republican Party. Thus, in analyzing voting distributions, the urban or rural status of a county becomes a critical factor, reflecting not only the geographical distribution of political preferences but also highlighting the political and social divides between urban and rural areas. 
Recognizing this spatial dependency helps in more accurately predicting election outcomes and in forming strategies for political parties.
Thus the first challenge lies in how to incorporate spatial information into transfer learning techniques.
\spacingset{1.5}%
\spacingset{1.9}%
Though the method proposed by \cite{tian2023transfer} provides a valuable framework to transfer learning and has been applied to the U.S. presidential election,  their method is for independent data and does not account for spatial information during transfer learning. 
Furthermore, the second challenge is the issue of data scarcity. It is evident that the average sample size per state (the number of counties) is only around 60. Relying only on data from swing states for spatial analysis could harm the accuracy of the estimates.

\paragraph{Data sources and processing:}%
The digital boundary definitions of the United States congressional district are available at \url{https://cdmaps.polisci.ucla.edu/}, and the cartographic boundary shape files of counties in the USA are available at the United States Census Bureau {(\url{https://www2.census.gov/geo/tiger/})}. 
The spatial weight matrix is constructed using above two geographic datasets. Here we apply the ``queen'' contiguity-based spatial weights and row-normalized. The ``queen'' criterion defines neighbors as spatial units sharing a common edge or a common vertex. The response is the difference of support at the county level for presidential candidates from both parties in 2016 and 2020, available at \url{https://github.com/tonmcg/US_County_Level_Election_Results_08-20}. The county-level demographic information forms the predictors, collected at \url{https://data.census.gov/}. 
Among all states and the federal district, we exclude Alaska, Hawaii, and Washington,~D.C\@.. In these 48 states, there are 3107 counties and we have 142 county-level predictors.  
% Then we add the pairwise interaction terms between predictors as new predictors. Finally, there are 1035 county-level predictors.

\subsection{Model Settings}\label{ssec: setups}
In order to deal with the two aforementioned challenges in U.S. presidential election prediction problems, we present a transfer learning framework within the spatial autoregressive (SAR) model.
For a pre-specified swing state as the target data,  we consider the following SAR model, 
\begin{equation}\label{mol: target model}
    Y^{(0)}=\sum_{l=1}^{p} \lambda_{l 0 } W_{l}^{(0)} Y^{(0)}+\sum_{j = 1}^{q} X^{(0)}_{j} \beta_{j 0}+V^{(0)},
\end{equation}
where the difference of support $Y^{(0)}$ is an $n_0\times 1$ vector of spatially correlated dependent variables, $W_{l}^{(0)}$ is an $n_0 \times n_0$ known non-stochastic spatial weight matrix for each $l \in [p]$, the county-level demographic information $X_{j}^{(0)}$ is an \(n_0 \times 1\) vector of exogenous regressors independent of the error term $V^{(0)}$ for each $j \in [q]$, \(V^{(0)}\) is an $n_0 \times 1$ vector of independently and identically distributed (i.i.d.) random variables with mean \(0\) and variance $\sigma^{2}>0$.
Let $\lambda_0 = {(\lambda_{1 0},\dots,\lambda_{p 0})}^{\trans}$, $\beta_0 = ( \beta_{10}, \dots,  \beta_{q 0})^{\trans}$ and \(\theta_0 \equiv \) \({({\lambda_0}^{\trans}, {\beta_0}^{\trans})}^{\trans}\).

Similar to general transfer learning settings as in~\cite{bastani2021predicting}, \cite{li2021targeting}, \cite{tian2023transfer}, in addition to the target study~\eqref{mol: target model}, additional samples are obtained from $K$ auxiliary studies (also termed as \textit{source datasets}) from other states, where \(K\) is assumed to be fixed. For the $k$-th source dataset, we also consider the SAR model,
\begin{equation}\label{mol: source model}
    \begin{aligned}
    Y^{(k)}= \sum_{l=1}^{p} \lambda_{l 0 }^{(k)} W_{l}^{(k)} Y^{(k)}
    +
    \sum_{j=1}^{q} X^{(k)}_{j} \beta^{(k)}_{j0}+V^{(k)}, \ k \in [K],
    \end{aligned}
\end{equation}
where \(Z^{(k)} \equiv (Y^{(k)}, X_{1}^{(k)},\dots, X_{q}^{(k)})\) is the $k$-th source dataset of sample size \(n_k\), $W_{l}^{(k)}$ is an $n_k \times n_k$ non-stochastic spatial weight matrix for each $l \in [p]$,
and $V^{(k)}$ is an $n_k \times 1$ vector of i.i.d. random variables with mean zero and variance $\sigma^{2}_k$, independent of the covariates \(X_{j}^{(k)}, j \in [q]\).
Let \(\lambda_{0}^{(k)} = {(\lambda_{1 0}^{(k)}, \dots, \lambda_{p 0 }^{(k)})}^{\T}\), $\beta_0^{(k)} ={(\beta_{1 0}^{(k)}, \dots, \beta_{q 0}^{(k)})}^{\T}$ and \(\omega^{(k)}_0 \equiv {({\lambda_{ 0 }^{(k)}}^{\T}, {\beta_0^{(k)}}^{\T})}^{\T}\) for the $k$-th source dataset.
Let \(\mathbf{X}^{(k)} \equiv (X_{1}^{(k)}, \dots, X_{q}^{(k)})\), \(\mathbf{W}^{(k)} \equiv ({W_{1}}^{(k)}, \ldots,{W_{p}}^{(k)})\), and \(\mathfrak{D}^{(k)} \equiv (Y^{(k)}, \mathbf{X}^{(k)}, \mathbf{W}^{(k)})\), \(k = 0, \dots, K\). In transfer learning, a similar auxiliary model to the target model is called an \textit{informative} model, and the ``similarity'' is characterized by   the difference between $\omega_0^{(k)}$ and $\theta_0$, i.e.\ \(\delta^{(k)} \equiv \) \(\theta_0 - \omega^{(k)}_0\). Given a tolerance level $h>0$, the index set of the informative auxiliary samples is defined as
\begin{equation} \label{eq: aux samples}
    \A \equiv \A_h =
    \{ 0 \leq k \leq K:  \| \delta^{(k)}\|_1 \leq h \},
\end{equation}
where \( \|\delta^{(k)}\|_1 \) is called the \textit{transferring level} of the \(k\)-th source dataset. In this context, the value of $h$ is used to balance the bias and the variance introduced by transfer learning. When $h$ is large, more labeled datasets are used, which reduces the prediction variance but increases the bias due to source datasets, and vice versa. We will discuss the value of $h$ in the conditions for theoretical results in Section~\ref{sec: theory}.%It is worth mentioning that the value of the tuning parameter $h$ remains unknown in general and it can be selected via some evaluation criteria such as cross-validation prediction errors in practice.

\section{Statistical Methodology}\label{sec: model}
% \blue{In this section, we introduce the SAR model and discuss transfer learning methods under the SAR model, including estimation with known informative source data and a transferable source detection algorithm. These procedures are summarized in Algorithm~\ref{algo1} and~\ref{algo3}.}
% \subsection{Model setting}

\subsection{Estimation with known informative source data}\label{ssec: known A}
In this section, we propose a two-step transfer learning procedure under the SAR model when the index set \(\A_h\) of the informative auxiliary samples is known, denoted as \textit{\( \A \)-TranSAR}.  In the first transferring step, we consolidate information from various sources by pooling all available auxiliary data, yielding a preliminary  estimator which is typically biased, since \(\omega^{(k)}_0 \neq \theta_0\) in general, although they are close. In the second debiasing step, we correct the estimation bias  by incorporating the target data via regularization.
The details of the two-stage estimation procedure are present in the following.

\paragraph{The transferring stage}
The first-step estimator, which is also called \textit{pre-training} model  in the transfer learning literature, with tuning parameter \(\lambda_{\omega}\) is defined as
\begin{equation}
    \label{est: A-tranSAR-step-one}
\wh{\omega}= \underset{\omega \in \mathbb{R}^{p+q}} {\arg \min }
\left\{
    \frac{1}{2n_{\mathcal{A}}}
    \sum_{k \in \mathcal{A}}
    \loss_1 \left(\mathfrak{D}^{(k)}\mid \omega \right)
    +\lambda_{\omega}
    \| \omega \|_1
\right\},
\end{equation}
where $\loss_1 ( \cdot \mid \omega)$ is a generic loss function with parameter \( \omega \), and \(n_{\mathcal{A}} \equiv \sum_{k\in \mathcal{A}} n_{k}\) is the total sample size of all informative source data. 
For the SAR model, the estimator could be the two-stage least squares (2SLS) estimator~\citep{kelejian1998generalized, lee2003best}, generalized method of moments estimator (GMM) \citep{kelejian1999generalized,lee2007gmm,liu2010gmm}, quasi-maximum likelihood estimator \citep[QMLE]{lee2004asymptotic} and so on. 
The main requirement for the first step estimator is a suitable convergence rate discussed in detail in Section~\ref{sec: theory}. 
Here the tuning parameter \(\lambda_{\omega}\) could be zero in low dimensional cases. When the dimension of parameters is greater than the sample size in high dimensional cases, the penalized SAR estimation with nonzero \(\lambda_{\omega}\) could produce a sparse estimator\citep{higgins2023shrinkage}. See also \citet{zhang2018spatial} for spatial weight matrices selection with model averaging, and \cite{tibshirani1996regression, fan2001variable, zhang2010nearly, wang2011random, su2017false} for more detail in regularization method.

\begin{remark}\label{rem: set A}

The index set \(\mathcal{A}\) may or may not contain the index $0$. This depends on the actual application scenarios.
In some cases the source data are available, and in the other cases, it may be difficult to obtain due to data privacy protection and security considerations.
In the latter case, the owner of the source data can provide the estimation results obtained by the first stage learner without the target data instead of the original source data, and the target task will use this result for the next debiasing stage.

\end{remark}

\paragraph{The debiasing stage}\label{sec: second step}

In the second step, the bias term estimator \(\wh{\delta}\) with tuning parameter \(\lambda_{\delta}\) using the target data is defined as
% \begin{equation}
%     \hat{\delta} = \underset{\delta \in \mathbb{R}^{p+q}}
%     {\arg \min }\left\{\frac{1}{2 n_{0}}
%     \loss_2 \left(\mathfrak{D}^{(0)} \mid \hat{\omega}\right) + \lambda_{\delta} \|\hat{\omega} - \delta \|_1
%     \right\},
% \end{equation}
% which is equivalent to minimize
\begin{equation}\label{est: A-tranSAR-step-two}
    \underset{\delta \in \mathbb{R}^{p+q}}
    {\arg \min } \frac{1}{2 n_{0}} \loss_2 \left(\mathfrak{D}^{(0)} \mid \wh{\omega}+ \delta \right) +\lambda_{\delta} \|\delta \|_1.
\end{equation}
Then, the bias-corrected \textit{\(\mathcal{A}\)-TranSAR} estimator is defined as
\begin{equation}
    \wh{\theta}_{\mathcal{A}\textit{-TranSAR}} = \wh{\omega} + \wh{\delta}.
\end{equation}

The two-stage  transfer learning procedure under the SAR model is summarized in Algorithm \ref{algo1}.
\spacingset{1}
\begin{algorithm}[htbp]
    \caption{\textit{\(\A\)-TranSAR}}\label{algo1} \KwIn{target dataset:
    \(\mathfrak{D}^{(0)}\); sources datasets: \((\mathfrak{D}^{(k)}, k\in \A)\);
    transferring set: \(\A\); and tuning parameters: \(\lambda_\omega\) and
    \(\lambda_\delta\)}

    \KwOut{\(\hat{\theta}_{\A\textit{-TranSAR}}\)}

    {
    \textbf{The transferring stage: }
    $$
    \wh{\omega}=
    \underset{\omega \in \mathbb{R}^{p+q}} {\arg \min }
    \left\{
        \frac{1}{2n_{\mathcal{A}}}
        \sum_{k \in \mathcal{A}}
        \loss_1 \left(\mathfrak{D}^{(k)}\mid \omega \right)
        +\lambda_{\omega}
        \| \omega \|_1
    \right\}
    $$\\
    }
    {
    \textbf{The debiasing stage: }
    $$
    \wh{\delta} = \underset{\delta \in \mathbb{R}^{p+q}}
    {\arg \min } \frac{1}{2 n_{0}} \loss_2 \left(\mathfrak{D}^{(0)} \mid \wh{\omega}+ \delta \right) +\lambda_{\delta} \|\delta \|_1.
    $$\\
    }
    {
    \textbf{Output:}
    \(\wh{\theta}_{\mathcal{A}\textit{-TranSAR}} =\wh{\omega} + \wh{\delta}\)
    }
\end{algorithm}
\spacingset{1.9}

\begin{remark}\label{rem:loss}
The loss functions \(\loss_1\) and \(\loss_2\) in the two stages are allowed to be different. 
In fact, we make no specific requirements on the form of \(\loss_1\) as long as the resulting estimator in the transferring stage satisfies the certain convergence rates discussed in Section 3.

In particular, we consider a classical 2SLS procedure in the second stage as an example, which has been widely used for the SAR model to deal with the endogeneity.
Given an \(n_0 \times d\) (\(d \geq p+q \)) instrumental variables matrix \(Q^{(0)}\) containing \(\mathbf{X}^{(0)}\), the first-stage estimator in the 2SLS procedure is
\begin{equation}\label{est: sls coef}
    \widehat{\pi}_j = \underset{\pi_j \in \mathbb{R}^{d}} {\arg \min } \left\{ \frac{1}{2n_0}\| {\mathbb{X}}^{(0)}_j - Q^{(0)}  \pi_j \|_2^2 \right\},
\end{equation}
for each \(j \in [p+q]\), where \({\mathbb{X}}^{(0)} = ({W_1}^{(0)} Y^{(0)}, \dots, {W_p}^{(0)} Y^{(0)}, \mathbf{X}^{(0)}) \in \mathbb{R}^{n_0\times(p+q)}\), \({\mathbb{X}}^{(0)}_j\) is the \(j\)-th column of \({\mathbb{X}}^{(0)}\) and $\pi_j$'s are coefficients of instrumental variables.
Denote the fitted values of \({\mathbb{X}}^{(0)}_j\)'s by \(\wh{\mathbb{X}}^{(0)}_j \equiv Q^{(0)} \hat{\pi}_j = P_{ Q^{(0)} } {\mathbb{X}}^{(0)}_j\), and \(\wh{\mathbb{X}}^{(0)} = (\wh{\mathbb{X}}^{(0)}_1, \dots, \wh{\mathbb{X}}^{(0)}_{p+q})\), where the projection matrix is $P_{Q^{(0)}} \equiv {Q^{(0)}} ( {Q^{(0)}}^\trans  Q^{(0)})^{-1} {Q^{(0)}}^\trans$. 
In the SAR model, instrumental variables could be 
\begin{equation*}
    Q^{(0)} =  (\mathbf{X}^{(0)}, W_1^{(0)} \mathbf{X}^{(0)} , \dots,  W_p^{(0)} \mathbf{X}^{(0)})
\end{equation*}
with \(d = q(1+p)\) columns. 
We note that if the first column of \(\mathbf{X}^{(0)}\) is the intercept term 1, we need to drop it in \(W_j^{(0)} \mathbf{X}^{(0)} (j \in [p])\) to avoid multicollinearity.
See~\cite{kelejian1998generalized} and~\cite{lee2003best} for several alternative choices of instruments.
The corresponding loss function for the second stage estimation is
\begin{equation}\label{obj: A-tranSAR-step-two}
    \loss_2 \left(\mathfrak{D}^{(0)} \mid \theta \right) = \|Y^{(0)} - \wh{\mathbb{X}}^{(0)} \theta \|_2^2.
\end{equation}
Then, the bias-corrected \textit{\(\mathcal{A}\)-TranSAR} estimator can be equivalently obtained by
\begin{equation}
    \wh{\theta}_{\mathcal{A}\textit{-TranSAR}} = \underset{\theta \in \mathbb{R}^{p+q}} {\arg \min} \frac{1}{2 n_{0}} \loss_2 \left(\mathfrak{D}^{(0)} \mid \theta \right) +\lambda_{\delta} \| \theta - \wh{\omega}\|_1.
\end{equation}

\end{remark}

\begin{remark}
    For high dimensional settings in the reduced form, the number of instruments is larger than the sample size.
    We could consider using a penalized 2SLS procedure as
    \begin{equation}\label{est: psls coef}
        \widehat{\pi}_j = \underset{\pi_j \in \mathbb{R}^{d}} {\arg \min } \left\{\frac{1}{2n_0}\| {\mathbb{X}}^{(0)}_j - Q^{(0)}  \pi_j \|_2^2 +p_{\lambda_{\pi_j} }(\pi_j) \right\},
    \end{equation}
    for certain penalty function \(p_{\lambda_{\pi_j} }(\cdot)\).
    In the literature,~\citet{belloni2012sparse} proposed the implementation of Lasso and post-Lasso methods to form the first-stage predictions and estimate optimal instruments in a linear instrumental variables model.
    \citet{fan2014endogeneity} developed a novel penalized GMM to cope with the endogeneity of a high dimensional model.
    \citet{cheng2015select} studied the problem of choosing valid and relevant moments for GMM estimation and developed asymptotic results for high-dimensional GMM shrinkage estimators that allow for non-smooth sample moments and weakly dependent  observations.
    \citet{zhu2018sparse} and~\citet{gold2020inference} introduced a two-stage Lasso procedure and established some finite-sample convergence rates under different assumptions.
\end{remark}

\subsection{A transferable source detection algorithm}\label{ssec: auto-detect}

The \textit{\(\mathcal{A}\)-TranSAR} estimator requires the true informative index set \(\mathcal{A}\) to be known correctly. However, in most empirical applications, the true informative set \(\mathcal{A}\) is unknown.
Transferring {adversarial auxiliary} samples may not improve the performance of the target model and could even cause worse results.
The effect of adversarial auxiliary samples is also called \textit{negative transfer}, meaning that certain inferior results are caused by uninformative source data \citep{torrey2010transfer,tian2023transfer}.

An intuitive approach might involve seeking a metric of similarity between  each source and the target data.
We compute the initial loss based on the target dataset. Subsequently, we combine the target dataset with each source dataset individually to estimate and compute the corresponding losses. If a source dataset is similar to the target dataset, we anticipate two loss values to be comparably close. Hence, we select a critical value to identify which source datasets are informative. For spatial datasets characterized by a network structure, data splitting used in \cite{tian2023transfer} can not be directly applied since it could disrupt this network integrity. 
As a result, we introduce the spatial residual bootstrap approach as an alternative.
This idea is illustrated in details as follows.

First, we compute an initial estimator \(\wh{\theta}_{ini} = {({\wh{\lambda}_{ini}}^{\trans}, {\wh{\beta}_{ini}}^{\trans})}^{\trans}\) using the target samples by certain estimation methods, such as 2SLS, QMLE or GMM, etc. Then, we obtain the residuals
$ \hat{e} = Y^{(0)} - \mathbb{X}^{(0)} \hat{\theta}_{ini},
$
where \({\mathbb{X}}^{(0)} = ({W}_1^{(0)}{Y}^{(0)}, \ldots, {W}_p^{(0)}{Y}^{(0)} , \mathbf{X}^{(0)})\).
Similar to residual bootstrap, we generate three independent copies, \(\wh{e}^{(1)}\), \( \wh{e}^{(2)}\) and \(\wh{e}^{(3)}\) of sample size \(n_0\) from the empirical distribution of \(\wh{e}\). Then, we generate three-fold bootstrap samples
\(Y^{(0,r)}\) and \( \mathbf{X}^{(0,r)} \) by
\begin{equation}
    \label{eq: Y and X gen}
\begin{aligned}
    Y^{(0,r)} = & \mathbb{X}^{(0)} \wh{\theta}_{ini} + \wh{e}^{(r)},~~
    \bX^{(0,r)} = & \mathbf{X}^{(0)},
\end{aligned}
\end{equation}
for \(r = 1, 2, 3\). Let \(Z^{(0,r)} = (Y^{(0,r)}, \mathbf{X}^{(0,r)})\) for
\(r = 1,2,3\). Let \(Z^{(0,-r)}\) be the concatenated bootstrap samples without the $r$-th sample, for example, $Z^{(0,-1)} = {({Z^{(0,2)}}^\trans, {Z^{(0,3)}}^\trans)}^\trans$.
Define \(\mathbf{X}^{(0, -r)}, \mathbb{X}^{(0, r)}\) and \(\mathbb{X}^{(0,
-r)}\) similarly.

Second, for each \(r = 1, 2, 3\), we fit the model in~\eqref{mol: target model} on the connected bootstrap samples \(Z^{(0,-r)}\) and use the resulting estimator \(\widehat{\theta}^{(0, r)}\)  to compute the loss value on the remaining bootstrap sample \(Z^{(0,r)}\), that is
\begin{equation}\label{eq: loss_r}
    \mathcal{L}^{[r]}(\widehat{\theta}^{(0, r)}) \equiv
    \frac{1}{2n_{0}}
    \| Y^{(0,r)} -  \mathbb{X}^{(0, r)} \widehat{\theta}^{(0, r)} \|_2^2.
\end{equation}
Then, we compute the average of  three loss values, \(\mathcal{L}^{(0)}=\sum_{r=1}^{3}\mathcal{L}^{[r]}(\widehat{\theta}^{(0, r)})/3\), which serves as the baseline loss value for the target dataset. We can also compute the standard deviation of
three loss values, denoted by
\begin{equation} \label{Eq: sigma}
    \hat{\sigma} =  \sqrt{\frac{1}{2}\sum_{r=1}^3{(\mathcal{L}^{[r]}
    (\hat{\theta}^{(0,r)}) - \mathcal{L}^{(0)})}^2},
\end{equation}
which measures the dispersion of bootstrapped loss values
$\mathcal{L}^{[r]}(\widehat{\theta}^{(0, r)})$ relative to their average $\mathcal{L}^{(0)}$.
Next, for each \(r = 1, 2, 3\), we combine \(Z^{(0,-r)}\) with each $k$-th source dataset, fit the model in  \eqref{mol: target model} on the newly combined data
and compute the loss denoted by \(\mathcal{L}^{(k,r)}\) on \(Z^{(0,r)}\). Similarly, the average loss for the $k$-th source dataset over three-fold bootstrap samples is \(\mathcal{L}^{(k)}\). Consequently, the difference between \(\mathcal{L}^{(k)}\) and \(\mathcal{L}^{(0)}\)
provides a metric of the similarity between the  $k$-th source dataset and the target data.
Source datasets whose corresponding loss differences below certain threshold are recognized as transferable informative data. Following the idea of \citet{tian2023transfer}, we choose $\hat{\sigma} \vee 0.01$ as the detection threshold. Then, the estimated transferable source index set \(\wh{\mathcal{A}}\) is defined as
\begin{equation} \label{eq: auto-detected A}
    \wh{\mathcal{A}} =
    \left\{
        k = 1, \ldots, K \mid \mathcal{L}^{(k)} - \mathcal{L}^{(0)} \leq  \left(\wh{\sigma} \vee 0.01 \right)
    \right\}.
\end{equation}
Once the transferable set $\mathcal{A}$ is estimated, we can apply the \textit{$\mathcal{A}$-TranSAR} in Algorithm \ref{algo1} to obtain the transfer learning estimator for the SAR model, denoted by  \(\wh{\theta}_{\wh{\mathcal{A}} \textit{-TranSAR}}\), which is termed as the \textit{TranSAR} estimator. The procedure described above is summarized in Algorithm~\ref{algo3}.

\spacingset{1}
\begin{algorithm}
    \caption{\textit{TranSAR}}\label{algo3}

    \KwIn{target: \(\mathfrak{D}^{(0)}\);
    all sources: \((\mathfrak{D}^{(k)}, k\in [K])\);
    tuning parameter: \(\lambda_{\theta}^{(k,r)}, k\in[K]\cup\{0\}, r \in
    [3]\); an initial estimator, \(\wh{\theta}_{ini}\).}

    \KwOut{$\wh{\theta}_\textit{TranSAR}$ and \(\wh{\A}\).}

    {
        Residuals: $\hat{e} = Y^{(0)} - \mathbb{X}^{(0)} \hat{\theta}_{ini}$ \;
    }
    \For{\(r\) from \(1\) to \(3\)}{
        Generate an independent copy \(\hat{e}^{(r)}\) from \(\hat{e}\)  and \(Y^{(0,r)}\), \(\bX^{(0,r)} \), \(Z^{(0,-r)}\),
        \(\mathbf{X}^{(0, -r)}, \mathbb{X}^{(0, r)}\) and \(\mathbb{X}^{(0,
        -r)}\)\;

        $\widehat{\theta}^{(0, r)} \leftarrow \underset{\theta \in
        \mathbb{R}^{p+q}} {\arg \min }{\frac{1}{2 n_{0}} \loss_2
        \left(Z^{(0,-r)} \mid \theta \right) +\lambda^{(0, r)}_{\theta} \|\theta
        \|_1}$ \;

        \For{\(k\) from \(1\) to \(K\)}{
            $
            \widehat{\theta}^{(k,r)} \leftarrow
            \underset{\theta \in \mathbb{R}^{p+q}} {\arg \min }\left\{\frac{1}{2\left(n_{k}+2n_{0}\right)}
            \left\{
            \loss_2 (\mathfrak{D}^{(k)} \cup Z^{(0,-r)} \mid \theta)
            \right\} +
            \lambda_{\theta}^{(k,r)}
            \| \theta \|_1
            \right\}
            $ \;
        }
    }

    Calculate the average losses: \(\mathcal{L}^{(k)} = \frac{1}{3}\sum_{r=1}^3
    \mathcal{L}^{[r]} (\hat{\theta}^{(k,r)}), k =0, 1, \dots, K\) \;

    Calculate the standard deviation: \(\hat{\sigma} \leftarrow
    \sqrt{\frac{1}{2}\sum_{r=1}^3{(\mathcal{L}^{[r]} (\hat{\theta}^{(0,r)}) -
    \mathcal{L}^{(0)})}^2}\) \;

    The estimated transferable source index set:
    \(\hat{\mathcal{A}} \leftarrow \left\{ k = 1, \ldots, K \mid 
    \mathcal{L}^{(k)} - \mathcal{L}^{(0)} \leq  \left(\hat{\sigma} \vee 0.01
    \right) \right\}\);

    \(\wh{\theta}_\textit{TranSAR}\)
    is obtained by Algorithm~\ref{algo1} via letting $\mathcal{A}=\hat{\mathcal{A}}$\;

    % {
    % \textbf{Output:}
    % $\wh{\theta}_\textit{TranSAR}$, and \(\wh{\A}\).
    % }
\end{algorithm}
\spacingset{1.9}

\section{Simulation}\label{sec: simulation}

In this section, we design several simulation experiments to investigate the performance of the proposed method (denoted as \textit{TranSAR}). We compare the finite sample performance of our estimator with the classical 2SLS estimation (which is fitted only using the target dataset, denoted as \textit{SAR}), the \textit{${\A}$-TranSAR} estimator (which is fitted based on the true informative index set $\A$, denoted as \textit{Oracle TranSAR}), pooled-estimator (which is fitted using all the source datasets as the transferable sets, denoted as \textit{$[K]$-TranSAR}), and the generalized linear model transfer learning algorithm (denoted as \textit{glmtrans}) proposed by \citet{tian2023transfer}.
The \textit{glmtrans} ignores the spatial information and can be considered as a baseline estimator to show that it is necessarily important to consider the spatial correlations if they exist.

We generate the data according to Models~\eqref{mol: target model} and~\eqref{mol: source model}. In these models, $n_{0}=256,n_{1}=n_{2}=\cdots=n_{K}=100,p=1,q=200,K=20$. The covariates $\mathbf{x}_{i}^{(k)}$ are generated by multivariate normal distribution with covariance matrix $\Sigma\in\mathbb{R}^{q\times q}$, whose the $(j,j')$-th entry is denoted as $\Sigma_{j,j'}$ for all $0\leq k\leq K$. We consider three designs for $\Sigma$: (1) $\Sigma_{j,j'}=1$ for $j=j'$ and $\Sigma_{j,j'}=0$ otherwise, (2) $\Sigma_{j,j'}=0.5^{|j-j'|}$, (3) $\Sigma_{j,j'}=0.9^{|j-j'|}$. We consider two different distributions for the i.i.d.\ error terms: (1) the standard normal distribution $N(0,1)$, and (2) the $t_{2}$ distribution.
The spatial units are assumed to be located on a square grid, and we set up several candidate spatial weight matrices as follows:
(1) the first \(W_{1}\) is the spatial weight matrix where spatial units interact with their neighbors on the left and right sides;
(2) the second \(W_2\) is the spatial weight matrix where spatial units interact with their neighbor above and below;
(3) the rest of the spatial weight matrices are assumed to be matrices where spatial units interact with their second-nearest, third-nearest, and so on, denoted as $W_{3},\ldots, W_{N}$, where $N<\sqrt{n_{0}}$.
In each dataset, we randomly draw spatial weight matrices from these candidate matrices. The weight matrices ${\{W_{k,l}\}}_{k\in\{0\}\cup[K],l\in[p]}$ are drawn randomly from $\{W_{1},\ldots,W_{N}\}$ without replacement. In Models~\eqref{mol: target model} and~\eqref{mol: source model}, we consider coefficients settings as follows. In the target Model~\eqref{mol: target model}, $\lambda_{0}$ $=0.4$ and $\beta_{0}$ $=(\mathbf{1}_{3}^{\trans},\mathbf{0}_{q-3}^{\trans})^{\trans}\in\mathbb{R}^{q}$. In Model~\eqref{mol: source model},
\begin{align*}
    &\lambda_0^{(k)} =\lambda_{0}\mathbbm{1}\{k\in\A\}-\lambda_{0}\mathbbm{1}\{k\in\A^{c}\}, \\
    &\beta_{j 0}^{(k)} = \beta_{j 0} - 0.05 \mathbbm{1}\{j \in \H_k, k \in \A \} - 2 \mathbbm{1}\{j \in \H_k, k \in \A^c \}, j =1, \dots, q,
\end{align*}
{where $\H_{k}$ is a random subset of $[q]$ with $|\H_{k}| = H$ if $k\in\A$ (i.e., \(H\) is the number of different coefficients in the informative set), or $|\H_{k}|=\frac{q}{2}$ if $k\notin\A$.
So, on the informative datasets, the differences of different \(\beta_{j0}^{(k)}\)'s  are 0.05; on uninformative datasets, the ones of different \(\beta_{j0}^{(k)}\)'s are 2.}

\begin{figure}[hp!]
    \centering
    \includegraphics[width=1\textwidth]{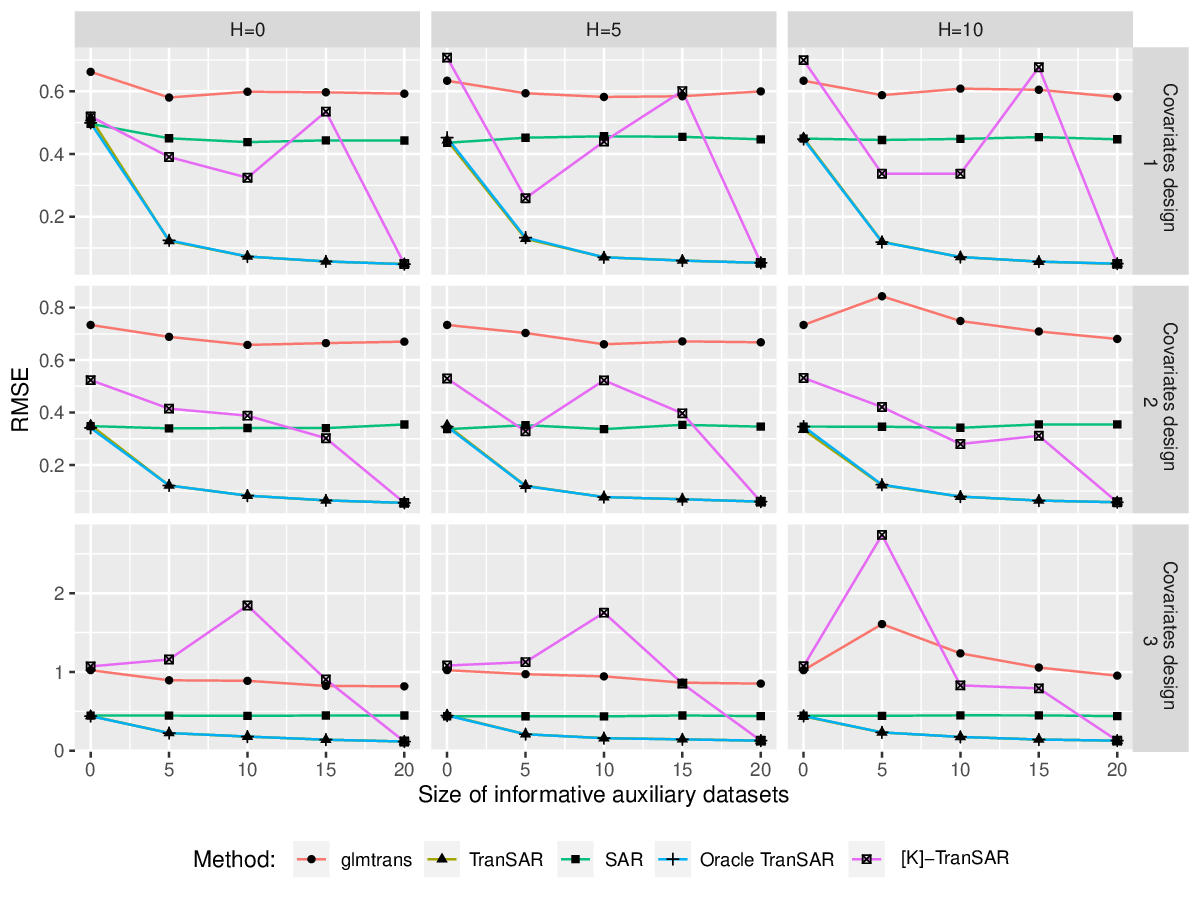}
    \caption{RMSE of Three Types of Covariates Designs with Normal Error Distribution, and $H = 0,5,10$. }\label{fig_sim_normal}
\end{figure}

We compute the root mean squared errors (RMSE),
\begin{equation}
\mbox{RMSE}=\sqrt{\frac{1}{R}\sum_{r = 1}^R  \| {\theta_0} - {\wh{\theta}}^{[r]} \|^2},
\end{equation}
where \( {\wh{\theta}}^{[r]}\) is the estimator of the \(r\)-th replication, based on the 2SLS loss with $q=200$
and display them under different settings in Figure~\ref{fig_sim_normal}.
In order to demonstrate the advantages of transfer learning, we let the size of informative auxiliary datasets $|\A|$ increase from $0$ to $20$ to show the decreasing trend in RMSE\@.
As the number of informative sets increases, compared to the classical 2SLS estimation and the \textit{glmtrans} method, we can see that \textit{$\wh{\A}$-TranSAR} and \textit{${\A}$-TranSAR} can significantly reduce the RMSE, resulting in more precise estimation.
Additionally, we can see that the \textit{$\wh{\A}$-TranSAR } performs almost the same as the oracle transfer learning method \textit{$\A$-TranSAR}.
That means, our proposed detection method is able to select informative sets with a high probability.
Moreover, from Figure~\ref{fig_sim_normal}, when the number of informative sets is small, the negative transfer effect of the pooled method, \textit{$[K]$-TranSAR} becomes apparent. Hence, it is necessary to select informative sets accurately. The same conclusion applies to all three types of design matrices and the three different settings for $H$. Detailed and comprehensive simulation results are compiled in Table~\ref{tab: sim result table} for covariates designs type 1.

\spacingset{1}

\begin{table}[hp!]
\centering
\resizebox{1\textwidth}{!}{
\begin{tabular}[t]{lllrrrrrr}
\toprule
\multicolumn{3}{c}{ } & \multicolumn{2}{c}{H=0} & \multicolumn{2}{c}{H=5} & \multicolumn{2}{c}{H=10} \\
\cmidrule(l{3pt}r{3pt}){4-5} \cmidrule(l{3pt}r{3pt}){6-7} \cmidrule(l{3pt}r{3pt}){8-9}
 & $|\mathcal{A}_h|$ & Method & RMSE of $\wh{\lambda}$ & Total RMSE & RMSE of $\wh{\lambda}$ & Total RMSE & RMSE of $\wh{\lambda}$ & Total RMSE\\
\midrule
\addlinespace[0.3em]
\multicolumn{9}{l}{\textbf{Covariates design  1}}\\
\hspace{1em} & $|\mathcal{A}_h|$ = 0 & SAR & 0.01931 & 0.49616 & 0.01593 & 0.44895 & 0.01507 & 0.43567\\

\hspace{1em} &  & $[K]$-TranSAR & 0.03309 & 0.51944 & 0.03755 & 0.69910 & 0.03892 & 0.70725\\

\hspace{1em} &  & glmtrans & NA & 0.66184 & NA & 0.63342 & NA & 0.63342\\

\hspace{1em} &  & TranSAR & 0.02088 & 0.51506 & 0.01574 & 0.45135 & 0.01574 & 0.44359\\

\hspace{1em} &  & Oracle TranSAR & 0.01938 & 0.49908 & 0.01556 & 0.44738 & 0.01638 & 0.45203\\

\hspace{1em} & $|\mathcal{A}_h|$ = 10 & SAR & 0.01536 & 0.43773 & 0.01568 & 0.44799 & 0.01633 & 0.45635\\

\hspace{1em} &  & $[K]$-TranSAR & 0.07954 & 0.32422 & 0.07143 & 0.33693 & 0.06139 & 0.43960\\

\hspace{1em} &  & glmtrans & NA & 0.59838 & NA & 0.60811 & NA & 0.58176\\

\hspace{1em} &  & TranSAR & 0.00868 & 0.07301 & 0.00842 & 0.07117 & 0.00842 & 0.07073\\

\hspace{1em} &  & Oracle TranSAR & 0.00873 & 0.07296 & 0.00835 & 0.07138 & 0.00844 & 0.07052\\

\hspace{1em} & $|\mathcal{A}_h|$ = 20 & SAR & 0.01552 & 0.44279 & 0.01598 & 0.44674 & 0.01568 & 0.44669\\

\hspace{1em} &  & $[K]$-TranSAR & 0.00643 & 0.04898 & 0.00692 & 0.04951 & 0.00614 & 0.05267\\

\hspace{1em} &  & glmtrans & NA & 0.59203 & NA & 0.58152 & NA & 0.59955\\

\hspace{1em} &  & TranSAR & 0.00648 & 0.04909 & 0.00693 & 0.04963 & 0.00613 & 0.05275\\

\hspace{1em} &  & Oracle TranSAR & 0.00642 & 0.04853 & 0.00698 & 0.05011 & 0.00614 & 0.05288\\
\cmidrule{1-9}
\addlinespace[0.3em]
\multicolumn{9}{l}{\textbf{Covariates design  2}}\\
\hspace{1em} & $|\mathcal{A}_h|$ = 0 & SAR & 0.01284 & 0.34807 & 0.01256 & 0.34623 & 0.01203 & 0.33645\\

\hspace{1em} &  & $[K]$-TranSAR & 0.03664 & 0.52357 & 0.03667 & 0.53178 & 0.03734 & 0.52947\\

\hspace{1em} &  & glmtrans & NA & 0.73390 & NA & 0.73390 & NA & 0.73390\\

\hspace{1em} &  & TranSAR & 0.01305 & 0.35195 & 0.01173 & 0.33473 & 0.01309 & 0.35113\\

\hspace{1em} &  & Oracle TranSAR & 0.01188 & 0.34144 & 0.01279 & 0.34645 & 0.01280 & 0.34603\\

\hspace{1em} & $|\mathcal{A}_h|$ = 10 & SAR & 0.01254 & 0.34102 & 0.01179 & 0.34173 & 0.01204 & 0.33671\\

\hspace{1em} &  & $[K]$-TranSAR & 0.02236 & 0.38791 & 0.02557 & 0.27970 & 0.03508 & 0.52258\\

\hspace{1em} &  & glmtrans & NA & 0.65766 & NA & 0.74894 & NA & 0.66023\\

\hspace{1em} &  & TranSAR & 0.00615 & 0.08292 & 0.00586 & 0.07913 & 0.00561 & 0.07733\\

\hspace{1em} &  & Oracle TranSAR & 0.00605 & 0.08292 & 0.00581 & 0.07938 & 0.00567 & 0.07754\\

\hspace{1em} & $|\mathcal{A}_h|$ = 20 & SAR & 0.01314 & 0.35434 & 0.01307 & 0.35456 & 0.01282 & 0.34617\\

\hspace{1em} &  & $[K]$-TranSAR & 0.00465 & 0.05552 & 0.00479 & 0.05768 & 0.00418 & 0.05992\\

\hspace{1em} &  & glmtrans & NA & 0.67016 & NA & 0.68051 & NA & 0.66777\\

\hspace{1em} &  & TranSAR & 0.00462 & 0.05539 & 0.00479 & 0.05770 & 0.00420 & 0.06011\\

\hspace{1em} &  & Oracle TranSAR & 0.00462 & 0.05542 & 0.00479 & 0.05763 & 0.00418 & 0.06007\\
\cmidrule{1-9}
\addlinespace[0.3em]
\multicolumn{9}{l}{\textbf{Covariates design  3}}\\
\hspace{1em} & $|\mathcal{A}_h|$ = 0 & SAR & 0.01044 & 0.44594 & 0.01052 & 0.44397 & 0.01006 & 0.43836\\

\hspace{1em} &  & $[K]$-TranSAR & 0.01525 & 1.07038 & 0.01479 & 1.07295 & 0.01464 & 1.08117\\

\hspace{1em} &  & glmtrans & NA & 1.02237 & NA & 1.02237 & NA & 1.02237\\

\hspace{1em} &  & TranSAR & 0.01000 & 0.44203 & 0.00959 & 0.43640 & 0.01076 & 0.45431\\

\hspace{1em} &  & Oracle TranSAR & 0.01002 & 0.43889 & 0.01011 & 0.44119 & 0.01053 & 0.44673\\

\hspace{1em} & $|\mathcal{A}_h|$ = 10 & SAR & 0.01070 & 0.44301 & 0.01078 & 0.44921 & 0.00960 & 0.43615\\

\hspace{1em} &  & $[K]$-TranSAR & 0.02097 & 1.84294 & 0.02474 & 0.82913 & 0.02935 & 1.75217\\

\hspace{1em} &  & glmtrans & NA & 0.88731 & NA & 1.23627 & NA & 0.94350\\

\hspace{1em} &  & TranSAR & 0.00508 & 0.17931 & 0.00512 & 0.17545 & 0.00493 & 0.15996\\

\hspace{1em} &  & Oracle TranSAR & 0.00508 & 0.17984 & 0.00507 & 0.17543 & 0.00493 & 0.15873\\

\hspace{1em} & $|\mathcal{A}_h|$ = 20 & SAR & 0.01058 & 0.44641 & 0.01009 & 0.43955 & 0.00990 & 0.43987\\

\hspace{1em} &  & $[K]$-TranSAR & 0.00381 & 0.11680 & 0.00413 & 0.12832 & 0.00340 & 0.12749\\

\hspace{1em} &  & glmtrans & NA & 0.81752 & NA & 0.95240 & NA & 0.85179\\

\hspace{1em} &  & TranSAR & 0.00383 & 0.11714 & 0.00413 & 0.12769 & 0.00340 & 0.12760\\

\hspace{1em} &  & Oracle TranSAR & 0.00382 & 0.11678 & 0.00413 & 0.12836 & 0.00341 & 0.12748\\
\bottomrule
\end{tabular}}

\caption{The RMSE of Various Estimators with Normal Distribution Error Terms. The RMSE of $\wh{\lambda}$ is the RMSE of the estimated coefficients of the $\lambda_0$. The total RMSE is the RMSE of all the estimated coefficients of the covariates. ``NA'' indicates that ``glmtrans'' is not applicable for the estimation of \(\lambda_0\).}\label{tab: sim result table}
\end{table}
\spacingset{1.9}

\section{An Application to U.S. Presidential Election Prediction}\label{sec: emp2}
In this section, we apply the proposed method in Algorithm~\ref{algo3} to empirical data from the U.S. presidential election.
As shown in Section~\ref{ssec: data},  the response variable \(Y^{(k)}\) is the difference of support rates in the county level, and  covariates \(\mathbf{X}^{(k)}\) include demographic and geographica data, such as including total population, total housing units, poverty people population, etc.
With the observed spatial weight matrix \( \mathbf{W}^{(k)}\), each swing state is regarded as a target with \(k=0\), and all the other states are considered as sources \(k \in [47]\). The target model~\eqref{mol: target model} and sources models~\eqref{mol: source model} could formulated as
\begin{equation*}
    \begin{aligned}
    Y^{(k)}= \sum_{l=1}^{p} \lambda_{l 0 }^{(k)} W_{l}^{(k)} Y^{(k)}
    +
    \sum_{j=1}^{q} X^{(k)}_{j} \beta^{(k)}_{j0}+V^{(k)}, \ k \in \{0\}\cup[K]. 
    \end{aligned}
\end{equation*}
We run the proposed estimation method, \textit{TranSAR}, using the polls from the 2016 U.S. presidential election and some other demographic data with geographic information.
We also estimate the SAR model with Lasso penalty (due to high dimensional covariates) using only the target datasets (denoted as SAR)  to compare the empirical performance.
Due to the inherent stochastic nature of the resampling procedure in the algorithm, we conduct multiple replications (20 times) for both methods to obtain a stable result.
The average prediction RMSE of the county-level vote results of the county-level 2020 U.S. presidential election is compared at Table~\ref{tab: emp compare 2016 using transar and sar}.
For all swing states, the proposed \textit{TranSAR} method performs better than the traditional SAR in terms of RMSE, which demonstrates the necessarity of transfer learning in the SAR framework.
\begin{table}[ht]
\centering
\begin{tabular}{lcccccc}
\toprule
& Florida & Georgia & Michigan & Minnesota & North Carolina & Ohio \\
\midrule
SAR & 0.10723 & 0.10995 & 0.10494 & 0.19211 & 0.10118 & 0.10984 \\
tranSAR & 0.09417 & 0.09064 & 0.08033 & 0.09646 & 0.10966 & 0.08297 \\
\bottomrule
\end{tabular}
\caption{Average Prediction RMSE of the Predicted County-level Rates of Support in the 2020 Presidential Election.}\label{tab: emp compare 2016 using transar and sar}
\end{table}

Moreover, we also consider the practical rules of the U.S. presidential election, which operate on a ``winner-takes-all'' basis.
In this context, the winner party secures all the state's electoral votes. 
Given the ``winner-take-all'' framework of the U.S. presidential election, we focus on predicting the winner party in each state.
Leveraging the county-wise population, we compute the final difference of support in the state-level endorsement for presidential contenders belonging to the respective parties.
We calculate the rate of support for state-level elections using the following formula,
{\small
\begin{equation*}
    \textrm{Rate of state-level support} =
    \sum_{i}
    \frac{\textrm{Rate of county \(i\) level support}\times \textrm{Votes of the county \(i\)}}{\textrm{Total votes}}.
\end{equation*}
}
Then, we compare the prediction rate of state-level support and the true one in $2020$ election, showing in Table~\ref{tab: emp compare 2016 using transar and sar in state level}. It demonstrates that the transfer learning under the SAR framework is able to effectively improve the state-level prediction.
\begin{table}[h]
    \centering
    \begin{tabular}{lcccccc}
        \toprule
        & Florida & Georgia & Michigan & Minnesota & North Carolina & Ohio \\
        \midrule
        SAR & -0.01989 & -0.08110 & -0.07157 & -0.04189 & -0.01782 & -0.05765 \\
        tranSAR & 0.00546 & -0.04186 & 0.03522  & -0.02228 & -0.02163  & -0.03189 \\
        \bottomrule
    \end{tabular}
    \caption{Prediction Bias of the Predicted State-level 2020 Election Results.}\label{tab: emp compare 2016 using transar and sar in state level}
\end{table}

The calculated ratios of correct predictions out of the 20 election result forecasts are collected in Table~\ref{tab: emp compare ele result}.
As shown in  Table~\ref{tab: emp compare ele result}, the method we proposed exhibits better performance, especially in predicting election outcomes in Michigan and Minnesota.
Simultaneously, it is noteworthy that for the election predictions in Georgia, neither of the methods works well.
It can be explained by the historical context: from $1972$ to $2016$, Georgia consistently supported the Republican Party, except for Democratic candidates from the southern region.
However, the state has become an increasingly competitive battleground.
In 2020, the Democratic candidate, Joe Biden, secured a victory over Donald Trump by a narrow margin of $0.2\%$.
This marks a significant departure from the history, as the Democrats secured the presidential election with a slim advantage. Evidently, Georgia has become an outlier in this regard. 
It is evident that predicting the ultimate election outcome for a single state is significantly more challenging than forecasting the rate of state-level support.
Nevertheless, we have made commendable progress in this regard.

\begin{table}[h]
\centering
\begin{tabular}{lcccccc}
    \toprule
    & Florida & Georgia & Michigan & Minnesota & North Carolina & Ohio \\
    \midrule
    SAR & 100\% & 0\% & 35\% & 90\% & 100\% & 100\% \\
    tranSAR & 100\% & 0\% & 100\% & 100\% & 100\% & 100\% \\
    \bottomrule
\end{tabular}
\caption{The Ratios of Correct Predictions for Election Result of 2020 Election Using the 2016 Election Data as Training Data.}\label{tab: emp compare ele result}
\end{table}

\begin{figure}[h!]
    \centering
    \resizebox{0.9\textwidth}{!}{
        \includegraphics[width=1\textwidth]{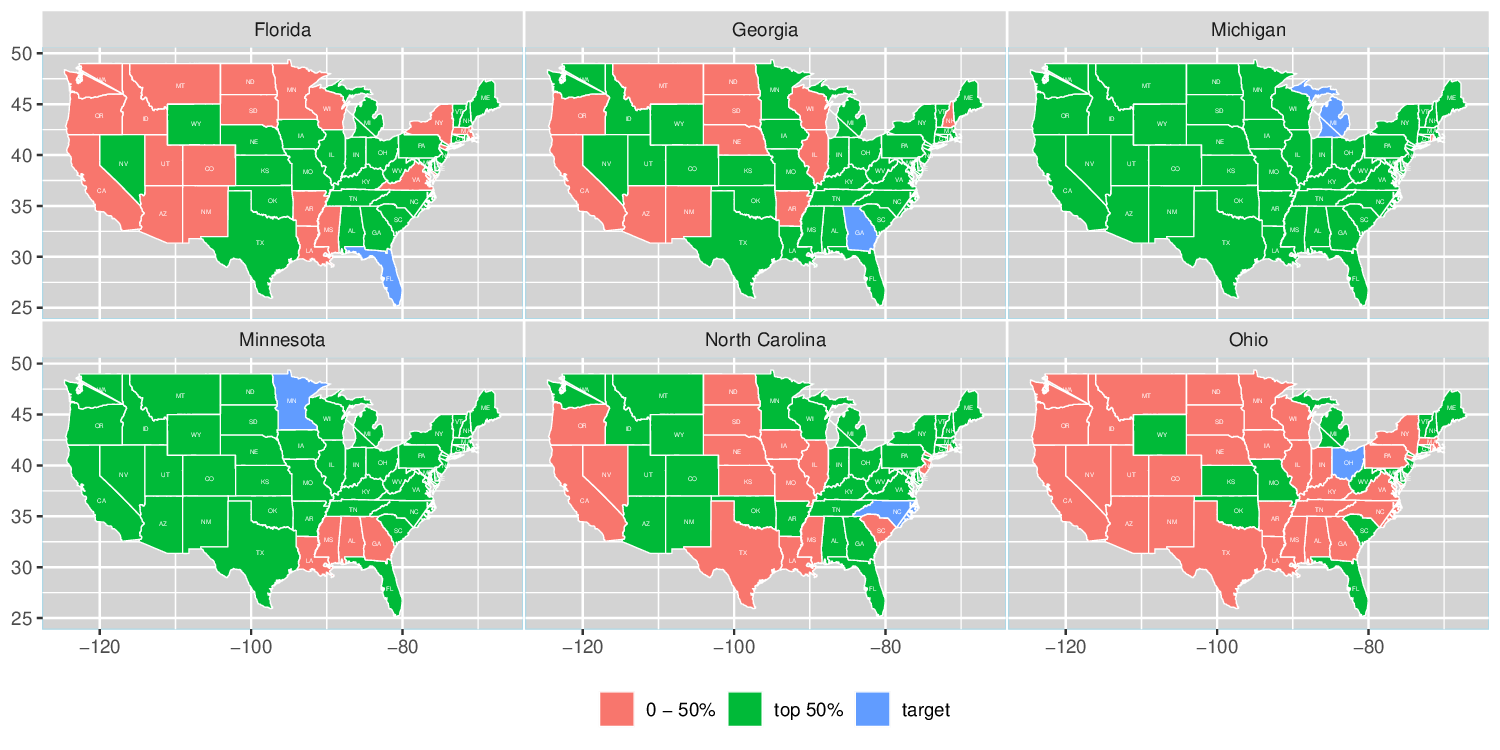}
        }
    \caption{The Detected Transferable States for Each Swing State.}\label{fig_emp_transferable_states_map_top_50}
\end{figure}

Next, we attempt to visualize transferable states, conducting a statistical analysis of the results after multiple repetitions. 
For a swing state, we record the frequency of occurrences for all transferable states and depict states with more than 50\% occurrences on the map in green color. As observed in Figure~\ref{fig_emp_transferable_states_map_top_50}, some transferable states exhibit geographical information signals. 
However, for some targets, the spatial or geographical correlations may not be readily apparent. 
This is a commonplace occurrence. 
With the development of the economy, culture, and scientific technology, regional interconnections are no longer confined solely to spatial proximity. 
Latent economic and political affiliations become pivotal factors.  

In addition, to provide a comprehensive demonstration of the superiority of the proposed TranSAR method, we treat each state as the target once a time. Subsequently, we conduct both the tranSAR and SAR procedures, and the outcomes have been summarized in Supplementary Materials. In conclusion, our tranSAR method consistently outperforms the traditional SAR method in the majority of cases, yielding superior prediction results.

\begin{figure}[ht!]
    \centering
    \includegraphics[width=0.85\linewidth]{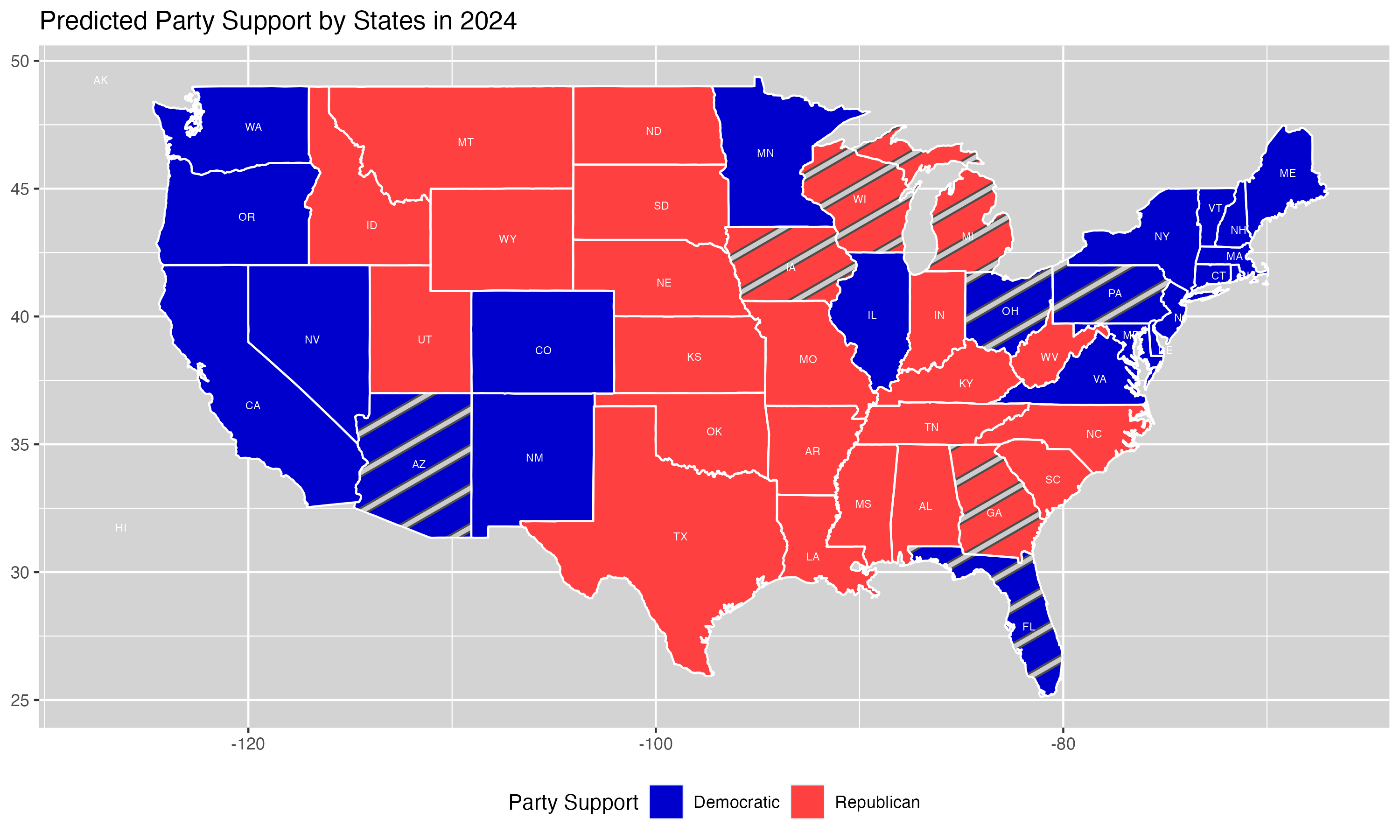}
    \caption{Predicted Party Support by States in 2024. The striped states are the swing states whose election outcomes predicted by  our TranSAR method.}
    \label{fig:map_pred_2024}
\end{figure}

Motivated by public interest, we also predict the outcome of the 2024 U.S. presidential election for all states using demographic and geographical data in 2022, based on the trained tranSAR model from the 2020 U.S. polling data along with other demographic and geographical data. We use the 2022 demographic data as covariates to predict county-level support rates. 
And we use the 2020 total vote counts of county-level as weights to calculate state-level support rates (because in the U.S., the ``winner-takes-all" system weights each county by its vote count).
We focus on the predictions of election outcomes for eight ``swing states'' following the rule based on whether there was a change in the supported party over the past three elections.
Therefore, we select {Arizona, Georgia, Florida, Pennsylvania, Michigan, Wisconsin, Ohio, and Iowa} as swing states. 
For non-swing states, we use the results from their last three presidential elections as the outcome for this election, while for swing states, we use our tranSAR method for prediction. Considering the randomness inherent in the algorithm, we conduct multiple rounds of analysis using our {TranSAR} procedure. In these analyses, if the Democratic Party receive support in more than half of the iterations, the swing state was classified as supporting the Democratic Party; otherwise, it was classified as supporting the Republican Party.
We obtain the supporting party for each state (see Figure~\ref{fig:map_pred_2024}) and calculate the final support vote counts combined with the electoral votes. The final prediction result shows that the Democratic Party receive {309} electoral votes, exceeding the threshold of 269 votes. Therefore, we predict that the outcome of the 2024 U.S. presidential election will favor {the Democratic Party}. 

% \orange{
% % media sources vary in their listings. For our analysis, we consider the list of swing states:\textbf{Arizona, Georgia, Michigan, Nevada, Pennsylvania, and Wisconsin}, as highlighted by sources from BBC, Forbes, and Axios.
% % The swing states show a noticeable split in their support. Arizona and Pennsylvania lean toward the Democratic Party, while Georgia, Michigan, Nevada, and Wisconsin are more inclined to support the Republican Party. Overall, the Republicans hold a slight advantage across these key states, though the Democrats maintain significant support in certain areas. The electoral votes from these battleground states will play a crucial role in determining the outcome of the election.
% }

\section{Theoretical results}\label{sec: theory}

In this section, we will establish an estimation convergence rate for the proposed \textit{$\mathcal{A}$-TranSAR} estimator by assuming $\underline{n}\equiv\min_{k\in[K]\cup\{0\}}n_{k}\rightarrow\infty$. Subsection \ref{sec: general theory} is for general results and Subsection \ref{sec: theory for ex} is based on the 2SLS estimation method. 
Finally, we prove the consistency of the transferable source detection algorithm in Subsection \ref{subsec:Detection-Consistency}.

For theoretical analysis, we define the population form for the above estimator as
\begin{equation}\label{pop: A-tranSAR-step-one}
    \omega^{\mathcal{A}}=\underset{\omega\in\mathbb{R}^{p+q}}{\arg\min}\sum_{k\in\mathcal{A}}\alpha_{k}\bar{\mathit{\ell}}_{1}^{(k)}(\omega),
\end{equation}
where $\alpha_{k}=\frac{n_{k}}{n_{\mathcal{A}}}$ and $\bar{\mathit{\ell}}_{1}^{(k)}(\omega)$ is the corresponding population objective function for dataset $\mathfrak{D}^{(k)}$. Under some regularity conditions, the first-stage estimator in Eq.~\eqref{est: A-tranSAR-step-one} is consistent, i.e., $\hat{\omega}-\omega^{\mathcal{A}}\xrightarrow{p}0$. Analogously,  define the ``true parameter'' based on the single source $k$ as
\begin{equation}\label{pop: A-tranSAR-step-one-k}
    \omega^{(k)}=\underset{\omega\in\mathbb{R}^{p+q}}{\arg\min}\ \bar{\mathit{\ell}}_{1}^{(k)}(\omega).
\end{equation}

\subsection{Convergence rate when \texorpdfstring{$\mathcal{A}_{h}$}{Ah} is known}\label{sec: general theory}

\begin{assumption}[Network structure]\label{assump: bounded matrix}
The spatial weight matrices $W_{l}^{(k)} (k\in \{0\}\cup [K], l\in [p])$ are non-stochastic with zero diagonals, and $p$ is fixed and does not depend on $n_{0}$ or $n_{k}$. $S^{(k)}(\lambda)=I_{n_{k}}-\sum_{l=1}^{p}\lambda_{l}^{(k)}W_{l}^{(k)}$ is invertible for all $\lambda\in\Lambda$, where the parameter space $\Lambda$ is compact. 
Matrices $W_{l}^{(k)}$ and $S^{(k)}(\lambda)^{-1}$ are uniformly bounded in both row and column sum norms for all $n_{k},k\in[K]\cup\{0\},l\in[p]$ and $\lambda\in\Lambda$. And for all large enough $n_{k}$, $\|W_{l}^{(k)}S^{(k)}(\lambda)^{-1}\|_{F}/\sqrt{n_{k}}$ is bounded away from zero and infinity uniformly over $k\in[K]\cup\{0\},l\in[p]$ and $\lambda\in\Lambda$.

\end{assumption}

\begin{assumption}[Regular conditions for the design matrix and error terms]\label{assump: design and error} \ 
\begin{enumerate}
    \item[(1)]  Denote $\mathbf{X}^{(0)}\equiv{(\mathbf{x}_{1}^{(0)},\dots\mathbf{x}_{n_{0}}^{(0)})}^{\T}$ and the covariance matrix of $\mathbf{x}_{i}^{(0)}$ for each $i$ is $\Sigma_{0}$. For any ${\beta}\in\mathbb{R}^{q},{\mathbf{x}_{i}^{(0)}}^{\trans}{\beta}\ (i\in[n_{0}])$, are {$i.i.d.$} $\kappa_{0}\|{\beta}\|_{2}$ sub-Gaussian with mean zero, and $0<\underline{\psi_{0}}=\lambda_{\min}(\Sigma_{0})\leq\lambda_{\max}(\Sigma_{0})=\overline{\psi_{0}}<\infty$.
    \item[(2)] The error terms $V_{i}^{(0)}\ (i\in[n_{0}])$ are $i.i.d.\ $sub-Gaussian with mean zero and variance $\sigma^{2}$.
    \item[(3)] As $n_{0}$ $\rightarrow$ $+\infty$, $\mathbb{P}\of{\lambda_{\min}(\frac{(\wh{\bbX}^{(0)})^{\trans}\wh{\bbX}^{(0)}}{n_{0}})\geq\phi_{0}}\rightarrow1$, where $\phi_{0}$ is a positive constant.
\end{enumerate}
\end{assumption}

\begin{assumption}[Convergence rates of the first step estimator]\label{assump: conv rate of 1th}

The first step estimator $\wh{\omega}$ minimizing~\eqref{est: A-tranSAR-step-one}
satisfies $\|\wh{\omega}-\omega^{\mathcal{A}}\|_{1}=\Op{(a_{n_{\A}}^{(1)})}$,
and $n_{\A}^{-1}\|{{\bbX}^{\A}}(\wh{\omega}-\omega^{\mathcal{A}})\|_{2}^{2}=\Op(a_{n_{\A}}^{(2)})$,
where $a_{n_{\A}}^{(1)}\rightarrow0$ and $a_{n_{\A}}^{(2)}\rightarrow0$ as $n_{\A}\rightarrow\infty$, and ${\mathbb{X}}^{\A}$ is stack of $\mathbb{X}^{(k)}$ for $k\in\A$.

\end{assumption}

% \begin{assumption}[Restricted eigenvalue condition]
%     \label{assump: res eig cond}
%     % For any strictly positive sequence \(a_{n_0} = o(1)\), \(\forall u \neq 0
%     % \in \mathbbm{R}^{p+q} \), we have \(\mathbb{P}
%     % (\inf_{u\in\mathcal{B}_1(a_{n_0}) \backslash \{0\}}
%     % \frac{u^{\trans}{{\mathbb{X}}^{(0)}}^\trans {\mathbb{X}}^{(0)} u} {n_0} \geq
%     % \phi_0\norm{u}{2}^2)\rightarrow 0\) as \(n_0 \rightarrow \infty\), where
%     % \(\phi_0\) is a positive constant.

% \end{assumption}

\begin{assumption}[Instruments]\label{assump: instruments for target}
When $n_{0}$ is large enough, the IV matrix 
\begin{equation}
    Q^{(0)}=\of{\bX^{(0)},W_{1}^{(0)}\bX^{(0)},\ldots,W_{q}^{(0)}\bX^{(0)}}
\end{equation}
satisfies $0<\psi_{*}^{(0)}\leq\frac{1}{\sqrt{n_{0}}}\|Q^{(0)}\|_{2}\leq{\psi^{(0)}}^{*}<\infty$
for some constants $\psi_{*}^{(0)}$ and ${\psi^{(0)}}^{*}$. There
exist constant coefficient vectors ${\pi}_{j}\in\mathbb{R}^{d}$ for
$j\in[p+q]$ such that $\max_{j=1,\dots,p+q}\|\widehat{\pi}_{j}-{\pi}_{j}\|_{2}=\Op(\sqrt{\log(d)/n_{0}})$
and $\|\pi_{j}\|_{2}^{2}$ is uniformly bounded over $j$ and $n_{0}$,
where $\wh{\pi}_{j}\ (j\in[p+q])$ are defined in Eq.~\eqref{est: sls coef}.
\end{assumption}

\begin{remark} Assumption~\ref{assump: bounded matrix} is 
standard in spatial econometrics. See \citet{kelejian1999generalized},~\citet{lee2004asymptotic},~\citet{lee2007gmm}
and~\citet{lee2014efficient}. Meanwhile, we assume the number of
spatial weight matrices $p$ is fixed, as 
either one or very few spatial weight matrices are often used in most empirical studies. For example, in our empirical application, $p=1$. Assumption~\ref{assump: design and error}
requires the predictors and error terms to be sub-Gaussian, which
is standard in the literature of transfer learning, such as~\citet{li2022transfer}
and~\citet{tian2023transfer}. The eigenvalue condition in Assumption~\ref{assump: design and error}
(iii) is widely employed in high-dimensional statistical literature
or can be derived from various regular conditions~\citep{mendelson2008uniform,negahban2009unified,raskutti2010restricted,vandegeer2009conditions,loh2012highdimensional}.
Assumption~\ref{assump: conv rate of 1th} imposes some conditions
on the convergence rate of the first step estimator, which will be
verified using a concrete example in Subsection~\ref{sec: theory for ex}.
Assumption~\ref{assump: instruments for target} is about the validity
of the instrument matrix $Q^{(0)}$ and the refitted value $\wh{\bbX}^{(0)}$. \end{remark}

\begin{theorem}\label{thm: error bound}
\textbf{\textsc{(Convergence rate of the \textit{\(\A\)-TranSAR} estimator)}}
Let the parameter space be $\Theta=\Lambda\times\left\{ {\beta}\mid\|{\beta}\|_{2}<M\right\} $ for some positive constant $M$. Under Assumptions 1-4, if the loss function $\loss_{1}$ is twice differentiable,
$\sup_{k\in\mathcal{A}}\|{\Sigma^{\mathcal{A}}}^{-1}\Sigma^{(k)}\|_{1}=\check{C}<\infty$,
$s\frac{\log{q}}{n_{0}}=o(1)$, $q\asymp d$, and $\lambda_{\delta}\gtrsim\sqrt{\log{(q)}/n_{0}}$,
where
\begin{align}
\Sigma^{(k)}\equiv & \int_{0}^{1}\frac{\partial^{2}}{\partial\theta\partial\theta^{\trans}}\bar{\mathit{\ell}}_{1}^{(k)}(\theta_{0}+t(\omega^{(k)}-\theta_{0}))dt,\ k\in\mathcal{A},\label{eq: sigma k}\\
\Sigma^{\mathcal{A}}\equiv & \sum_{k\in\mathcal{A}}\int_{0}^{1}\alpha_{k}\frac{\partial^{2}}{\partial\theta\partial\theta^{\trans}}\bar{\mathit{\ell}}_{1}^{(k)}(\theta_{0}+t(\omega^{\mathcal{A}}-\theta_{0}))dt,\alpha_{k}=\frac{n_{k}}{n_{\mathcal{A}}}. \label{eq: sigma A}
\end{align}
Denote 
$S=\operatorname{support}(\theta_{0})$ and $s=|S|$. Then
\begin{equation}
\|\wh{\theta}-\theta_{0}\|_{2}^{2}=\Op\of{(\sqrt{\frac{\log{q}}{n_{0}}}h)\wedge(s\frac{\log{q}}{n_{0}})\wedge h^{2}+a_{n_{\A}}^{(2)}}.
\end{equation}
\end{theorem}

\begin{remark} When we consider the case of fixed dimension of covariates
$q$, under some regularity conditions, $a_{n_{\A}}^{(2)}=\frac{1}{n_{\A}}$.
Then, we could obtain $\|\wh{\theta}-\theta_{0}\|_{2}^{2}=\Op(\min(n_{0}^{-1},h^{2})+n_{\A}^{-1})$, which implies that when the sample size $n_{\A}$ of the informative source data is large enough and $h$ has a suitable convergence rate, the estimation convergence rate of the \textit{\(\A\)-TranSAR} estimator is better than that only using the target data.
\end{remark}

\subsection{Convergence rate under the 2SLS loss}\label{sec: theory for ex}

In this section, we verify the requirements in the above theorem for the 2SLS estimation
and simplify the convergence rate of the \textit{\(\A\)-TranSAR} estimator. 
As in Eq.~\eqref{obj: A-tranSAR-step-two}, let $\loss_{1}(\cdot\mid\omega)$ be the loss function of the 2SLS.
To elaborate, let $Y^{\mathcal{A}}$ denote the stack consisting of $Y^{(k)}$ for all $k\in\mathcal{A}$, i.e., $Y^{\mathcal{A}}={({Y^{(k)}}^{\trans},k\in\mathcal{A})}^{\trans}$.
$\mathbf{X}^{\mathcal{A}}$ and ${\mathbb{X}}^{\A}$ are defined similarly.
Then the objective function~\eqref{est: A-tranSAR-step-one} could be rewritten as
\begin{equation}
\frac{1}{2n_{\mathcal{A}}}\|Y^{\mathcal{A}}-\widehat{\mathbb{X}}^{\mathcal{A}}\omega\|_{2}^{2}+\lambda_{\omega}\|\omega\|_{1},
\end{equation}
where $\wh{\bbX}^{(k)}$ is defined similarly as $\wh{\bbX}^{(0)}$
in Subsection~\ref{sec: second step}, and $\wh{\bbX}^{\A}$ is their
stack. Here the IV matrices are $Q^{(k)}$ and $Q^{\mathcal{A}}$.
We impose several similar conditions on $\mathfrak{D}^{(k)}$
($k\in\mathcal{A}$) in the following.

\begin{assumption}\label{assump: design and error on aux k}
\textbf{\textsc{(Regular conditions for design matrices and error
terms on the $k$-th source)}}
For source $k\in\mathcal{A}$, we have the following assumptions:
(i) For the design matrix $\bX^{(k)}=(\mathbf{x}_{1}^{(k)},\dots\mathbf{x}_{n_{k}}^{(k)})^{\trans}$
and any ${\beta}\in\mathbb{R}^{q},{\mathbf{x}_{i}^{(k)}}^{\trans}{\beta}\ (i\in[n_{k}])$
are $i.i.d. \kappa_{k}\|{\beta}\|_{2}$ sub-Gaussian with mean
zero for certain constant $\kappa_{k}$, where $\max_{k\in\A}\kappa_{k}<\infty$.
(ii) The error terms $V_{i}^{(k)}\ (i\in[n_{k}])$ are {$i.i.d.$} sub-Gaussian
with mean zero and variance $\sigma_{k}^{2}$.
(iii) As $n_{k}\rightarrow+\infty$, $\mathbb{P}\of{\lambda_{\min}\of{\frac{{{\wh{\bbX}}^{(k)\T}}{\wh{\bbX}}^{(k)}}{n_{k}}}\geq\phi_{k}}\rightarrow1$,
where $\phi_{k}$ is a positive constant.
\end{assumption}

\begin{assumption}[Instruments for the $k$-th source]\label{assump: instruments}
For the instruments
\begin{equation}
Q^{(k)}=\of{\bX^{(k)},W_{1}^{(k)}\bX^{(k)},\ldots,W_{q}^{(k)}\bX^{(k)}},\ k\in\A\cup\{0\},
\end{equation}
when $n_{k}$, the number of rows of $Q^{(k)}$, is large enough,
$0<\psi_{*}^{(k)}\leq\frac{1}{\sqrt{n_{k}}}\|Q^{(k)}\|_{2}\leq{\psi^{(k)}}^{*}<\infty$
for some constants $\psi_{*}^{(k)}$ and ${\psi^{(k)}}^{*}$. There
exist constant coefficient vectors $\pi_{j}^{(k)}\in\mathbb{R}^{d}$
for $j\in[p+q]$ such that $\max_{j=1,\dots,p+q}\|\hat{\pi}_{j}-\pi_{j}^{(k)}\|_{2}=\Op(\sqrt{\log(d)/n_{k}})$ and $\|\pi_{j}^{(k)}\|_{2}$ is uniformly bounded over $j$ and $k$.

% \red{To consider: For any \({{\gamma}^{(k)}} \in \mathbb{R}^d, {Q^{(k)}_{(i,.)}}
% {{\gamma}^{(k)}} \  (i \in [n_0])\) are {$i.i.d. $}
% \(\kappa_0\|{\gamma}^{(k)}\|_2\) sub-Gaussians with mean zero. }
\end{assumption}

Under Assumptions~\ref{assump: design and error on aux k} and~\ref{assump: instruments},
the convergence rate of
the \textit{\(\A\)-TranSAR} estimator is summarized in the following Theorem~\ref{thm: error bound of ex1}.

\begin{theorem}\label{thm: error bound of ex1}

If Assumptions~\ref{assump: bounded matrix},~\ref{assump: design and error},~\ref{assump: instruments for target},~\ref{assump: design and error on aux k},
and~\ref{assump: instruments} hold, $\sup_{n_{k}}{\mathbf{B}^{(k)}}<\infty$
with ${\mathbf{B}^{(k)}}\equiv1\vee\|\beta_{0}^{(k)}\|_{2}$, $K$
is a fixed, $q\asymp d$, $s\frac{\log{q}}{n_{0}}=o(1)$, $s^{2}\frac{\log{n_{0}}}{n_{0}}=o(1)$,
$n_{\A}>n_{0}$, $q^{2}h=O(1)$, $\lambda_{\delta}\gtrsim\sqrt{(\log{q})/n_{0}}$
and $\lambda_{\omega}\gtrsim\sqrt{(\log{q})/n_{\A}}$, where $S=\operatorname{support}(\theta_{0})$
and $s=|S|$, then we have
\begin{equation}
\|\wh{\theta}-\theta_{0}\|_{2}^{2}=\Op\of{(\sqrt{\frac{\log{q}}{n_{0}}}h)\wedge({\frac{s\log{q}}{n_{0}}})\wedge h^{2}+(\frac{\log{n_{A}}}{n_{\A}}h^{2})\wedge\frac{\log^{3}{n_{A}}}{n_{\A}^{3}}
+\frac{s\log{q}}{n_{\A}}}.\label{eq: bound for thetahat with ex1}
\end{equation}
\end{theorem}
Theorem \ref{thm: error bound of ex1} indicates that if $h\ll s\sqrt{\log q/n_0}$ and $n_{\A}\gg n_0$,
the \textit{\(\A\)-TranSAR} estimator is better than
the classical 2SLS estimator based on only
the target data, which demonstrates the usefulness of transfer learning in the SAR models.

\subsection{Detection consistency}\label{subsec:Detection-Consistency}

Finally, we will show that our proposed transferable source detection
algorithm is consistent even for a high dimensional model, i.e., $q\gg n_{0}$.

\begin{assumption}[Identifiability condition]\label{assump: identifiability}

The tolerance level $h=o(1)$ as the minimal sample size $\underline{n}\equiv\min_{k\in[K]\cup\{0\}}n_{k}\rightarrow\infty$.
For any $k\in[K]\cup\{0\}$ and $r\in[3]$, $q\log(q)\|\widetilde{\theta}^{(k)}-\widehat{\theta}^{(k,r)}\|_{1}=o_{\mathbb{P}}(1)$.
For $k\in\mathcal{A}^{c}\equiv[K]-\mathcal{A}$, we control the signal power as
\begin{align}\label{cond: ic.2}
\underline{\lambda}\|\theta_{0}-\widetilde{\theta}^{(k)}\|_{2}^{2}\geq & \|{\theta}_{0}-\widetilde{\theta}^{(k)}\|_{1}+\bar{C}\left[(\|\theta_{0}-\widehat{\theta}^{(0,r)}\|_{1})\vee1\right]-\nonumber \\
 & \frac{\sigma^{2}}{2n_{0}}\operatorname{tr}\{[(S(\tilde{\lambda}^{(k)})-S)S^{-1}]^{\trans}[(S(\tilde{\lambda}^{(k)})+S)S^{-1}]\},
\end{align}
and the gap between $\theta_{0}$ and $\widetilde{\theta}^{(k)}$,
\begin{equation}
\varphi_{k}\equiv\sqrt{\frac{1}{n_{0}}}(\|\theta_{0}-\widetilde{\theta}^{(k)}\|_{1}+1)^{2}=o(1),\label{cond: ic.4}
\end{equation}
where $\bar{C}>0$ is a constant and $\underline{\lambda}\equiv\liminf_{n_{0}\rightarrow\infty}\lambda_{\min}(\frac{1}{2n_{0}}\mathbb{E}{{\bX}^{\star}}^{\T}{\bX}^{\star})>0$, with ${\bX}^{\star}\equiv(\mathbf{W}^{(0)}(\mathbf{I}_{p}\otimes(S^{-1}{\bX}^{(0)}{\beta}_{0})),{\bX}^{(0)})$.
\end{assumption} 
\begin{remark} 
    Conditions~\eqref{cond: ic.2} and~\eqref{cond: ic.4} guarantee a suitable gap between the population-level coefficients in set $\A^{c}$ and the true coefficient of the target model. Intuitively, assuming an appropriate signal strength is necessary to ensure the consistency of estimators. On the one hand, a very weak signal makes it challenging for the algorithm to differentiate informative and uninformative sources. On the other hand, an excessively strong signal can lead the first step estimator to mimic the model structure of the source, thereby hindering accurate identification. 
\end{remark}

Next, we establish the detection consistency of $\wh \A$ in the following theorem.

\begin{theorem}[Detection consistency of \(\wh \A\)]\label{thm: detection consistency}
If Assumptions~\eqref{assump: bounded matrix}-\eqref{assump: instruments for target} and~\eqref{assump: identifiability} are satisfied, $\|\beta_{0}\|_{2}<\infty$ and $\frac{\log q}{n_{0}}=o(1)$, then the automatically detected transferable index set $\wh{\A}$ in Eq.~\eqref{eq: auto-detected A} is consistent, i.e.,
\begin{equation}
\mathbb{P}(\wh{\A}=\A)\rightarrow1\text{, as }\underline{n}\rightarrow\infty.
\end{equation}
 \end{theorem}

\section{Conclusion}\label{sec: conclusion}
To more accurately explore the spatial factors influencing swing states, we propose a method to enhance the estimation of spatial model in swing states by transferring knowledge from other states. We introduce a transfer learning method within the framework of SAR models, aimed at addressing spatially dependent data of small sample size. The empirical application of this methodology, specifically in predicting U.S. election results in swing states using demographic and geographical data, demonstrates its potential to enhance traditional spatial methods. Based on our comprehensive analysis and the application of the tranSAR model, our prediction for the 2024 U.S. presidential election is a victory for the \textbf{Democratic} party.
Future research could focus on developing tests to detect informative sets and improving the accuracy of coefficient estimation within the transfer learning framework.

\section*{Supplementary Materials}

\noindent All technical proofs of theorems and the additional simulations are included in the online supplementary materials.

\if1\blind
{
\section*{Funding}
This work is supported by National Key R\&D Program of China(2022YFA1003800), the National Natural Science Foundation of China(NNSFC) (71988101, 72073110, 72333001, 12231011), National Statistical Science Research Grants of China(Major Program 2022LD08), and the 111 Project(B13028).
}\fi 

% \section*{Declaration of generative AI and AI-assisted technologies in the writing process}
% During the preparation of this work the authors used ChatGPT 4.0 in order to polish the writing. After using this tool/service, the authors reviewed and edited the content as needed and take full responsibility for the content of the publication.

% \bigskip
% \begin{center}
% {\large\bf SUPPLEMENTARY MATERIAL}
% \end{center}

% % %%bib page

\vskip 0.2in
\spacingset{1}
\bibliographystyle{apalike} %apa7th
\bibliography{tranSAR.bib}

\end{document}